\documentclass[bookmarks=true]{article}
\usepackage{amssymb}
\usepackage{amsmath}
\usepackage[final]{corl_2025} 
\usepackage{float}
\usepackage{afterpage}
\usepackage{stfloats}
\usepackage{tabularx}
\usepackage{array}
\usepackage{makecell}
\usepackage{graphicx}
\usepackage{booktabs}
\usepackage{multirow}
\usepackage{multicol}
\usepackage{dsfont}
\usepackage{hyperref}
\usepackage[dvipsnames]{xcolor}
\usepackage{xspace}
\usepackage{cite}
\usepackage{cuted}
\usepackage{caption}
\usepackage{bbm}
\usepackage{paralist}
\usepackage{enumitem}
\usepackage{mdframed} 
\usepackage[T1]{fontenc} 
\usepackage{algorithm}
\usepackage{algpseudocode}
\usepackage{times}
\usepackage{threeparttable}
\usepackage{hyperref}
\usepackage{colortbl}
\usepackage{wrapfig}
\usepackage{appendix}
\usepackage{tcolorbox}
\title{O$^3$Afford: One-Shot 3D Object-to-Object Affordance Grounding for Generalizable Robotic Manipulation}

\author{
    \textbf{Tongxuan Tian$^{1}$} \quad
    \textbf{Xuhui Kang$^{1}$} \quad
    \textbf{Yen-Ling Kuo$^{1}$} \quad 
    \vspace{3pt} \\
    $^{1}$University of Virginia   \\
    \texttt{\{nua3jz,xuhui,ylkuo\}@virginia.edu}
} 

\renewcommand{\eqref}[1]{Eqn~\ref{#1}}
\newcommand{\figref}[1]{Figure~\ref{#1}}

\newcommand{\tabref}[1]{Table~\ref{#1}}

\newcommand{\appendref}[1]{Appendix~\ref{#1}}

\def\rekep{ReKep\xspace}

\def\weblink{\urllink[pre = \bgroup\bf, post = \egroup]}

  {\begin{mdframed}[topline=false, bottomline=false, rightline=false,%
    linewidth=1pt,linecolor=black,%
    leftmargin=1cm,rightmargin=1cm,%
    innerleftmargin=10pt,innerrightmargin=10pt,%
    innertopmargin=0pt,innerbottommargin=0pt,%
    skipabove=\baselineskip,skipbelow=\baselineskip]%
   \fontfamily{ppl}\selectfont
  }%
  {\end{mdframed}}

\begin{document}
\maketitle

\vspace{-0.5cm}
\begin{abstract}
Grounding object affordance is fundamental to robotic manipulation as it establishes the critical link between perception and action among interacting objects. However, prior works predominantly focus on predicting single-object affordance, overlooking the fact that most real-world interactions involve relationships between pairs of objects. In this work, we address the challenge of object-to-object affordance grounding under limited data contraints. Inspired by recent advances in few-shot learning with 2D vision foundation models, we propose a novel one-shot 3D object-to-object affordance learning approach for robotic manipulation. Semantic features from vision foundation models combined with point cloud representation for geometric understanding enable our one-shot learning pipeline to generalize effectively to novel objects and categories. We further integrate our 3D affordance representation with large language models (LLMs) for robotics manipulation, significantly enhancing LLMs' capability to comprehend and reason about object interactions when generating task-specific constraint functions. Our experiments on 3D object-to-object affordance grounding and robotic manipulation demonstrate that our O$^3$Afford significantly outperforms existing baselines in terms of both accuracy and generalization capability. 
Project website: \href{https://o3afford.github.io/}{https://o3afford.github.io}.
\end{abstract}
\section{Introduction}
Affordance grounding enables machines to identify the functional properties of objects or environments that indicate potential interactions.
It has been used to effectively communicate how objects can be interacted with in numerous downstream tasks, including human-computer interaction~\citep{kaptelinin2012affordances}, visual understanding~\citep{delitzas2024scenefun3d}, and robotic manipulation~\citep{kuang2024ram}.
Many prior works have investigated affordance prediction in 2D pixel space~\citep{li2024one, luo2022learning, luo2023leverage, zhai2022one, zhao2020object}. However, these works predominantly focus on single-object affordance and neglect that many daily tasks involve object-to-object interactions.

Object-to-object affordance grounding addresses functional relationships in tasks that require meaningful interactions between pairs of objects. For instance, cutting typically requires two objects: a source (e.g., a knife) and a target (e.g., an apple) to interact in a specific spatial relation. A significant challenge here is the scarcity of annotated data for model training.
It is hard to collect and annotate large-scale data for object-to-object affordance grounding.
O2O-Afford~\citep{mo2022o2o} addresses this challenge in an annotation-free manner through automatic extraction of contacts in simulation, which renders it only applicable to simple affordances such as placing and fitting. Our work aims to infer complex affordances such as pouring and cutting, which are crucial in robotic manipulation.

We seek to develop a solution for generalizable object-to-object affordance grounding with minimal supervision. Recent advances in vision foundation models (VFMs)~\citep{oquab2023dinov2, pmlr-v139-radford21a, rombach2022high} have demonstrated strong zero-shot and few-shot generalization capabilities across various downstream tasks~\citep{radford2021learning,wang2024d3fields,zhou2022zegclip,Han_2024_CVPR,zhang2023tale}.
A recent study~\citep{li2024one} has demonstrated VFMs' potential for 2D affordance prediction; however, the limited geometric information available in images restricts generalization across different viewpoints and object categories, hindering practical application in robotic manipulation. We hypothesize that the geometric information provided by 3D representations such as point clouds, when combined with semantic features from VFMs, can enable robust generalization across varied geometries, unseen object instances, and entirely novel object categories.

In this paper, we introduce O$^3$Afford, a \textbf{O}ne-shot \textbf{O}bject-to-\textbf{O}bject \textbf{Afford}ance Grounding framework with 3D semantic point cloud distilled from VFMs. We first project features from DINOv2~\citep{oquab2023dinov2} from multi-view RGB-D observations onto the point clouds of both the source object and the target object to be manipulated. The two object point clouds, enriched with part-aware semantic features, are then processed through our bi-directional affordance discovery module, which considers the geometric contexts of both objects in both directions to predict the final 3D affordance map. We integrate the output of our affordance grounding module with large language models (LLMs) to generate constraints for optimization-based robotic manipulation, enabling enhanced spatial understanding capabilities compared to reasoning solely from images or point clouds in robotic manipulation.

This paper makes the following contributions:
1) We introduce O$^3$Afford, a method for learning object-to-object affordance in 3D with one-shot examples.
2) We effectively integrate object-to-object affordance into optimization-based robotics manipulation with constraints generated by LLMs.
3) We show that O$^3$Afford outperforms baselines of affordance prediction and robotic manipulation across a range of simulated and real-world robotic manipulation tasks, exhibiting strong semantic and geometric generalization capabilities.
\section{Related Work}

\textbf{Affordance Grounding.}
Several prior works have focused on learning 2D affordance, with established datasets and benchmarks~\citep{luo2022learning, zhai2022one, myers2015affordance, sawatzky2017weakly}.
The primary goal is to predict a 2D functional affordance map under a finite set of affordance types~\citep{luo2023leverage, jang2024intra, li2023locate} or language-conditioned settings~\citep{qian2024affordancellm, chen2024worldafford}.
Recent efforts have extended it to weak supervision~\citep{li2023locate, jang2024intra}, intra-class generalization~\citep{li2024one, ju2024robo}, and open-vocabulary~\citep{tong2024oval, qian2024affordancellm} settings.
However, limited geometric information in 2D images poses challenges for downstream tasks like spatial understanding and manipulation, motivating efforts to ground affordance in 3D point cloud~\citep{Deng_2021_CVPR}, from 2D images~\citep{Yang_2023_ICCV, yang2024lemon}, or language models~\citep{chu20253d, Li_2024_CVPR}.
While effective, these works neglect the reality that many daily tasks involve object-to-object interactions in which the geometry of both objects should be considered.
Furthermore, the challenge in this setting is the limited annotated data for training.
O2O-Afford~\citep{mo2022o2o} addresses this issue through automatic annotation by extracting contact information from randomized configurations in simulated environments; however, this approach limits its scalability and generalizability to broader real-world scenarios.
In comparison, our work aims to use minimal annotation (i.e., one-shot) for supervision while achieving better generalization, making it more suitable for real-world robotic manipulations.

\textbf{Few-shot Learning with Foundation Models.}
Foundation models have demonstrated remarkable capabilities as few-shot learners~\citep{NEURIPS2020_1457c0d6, song2022clip}. 
Pretraining on internet-scale datasets has endowed these models with common sense knowledge~\citep{pmlr-v139-radford21a, oquab2023dinov2, touvron2023llama, rombach2021highresolution}, enabling them to adapt to new tasks with a limited number of demonstrations.
ZegClip~\citep{zhou2022zegclip} extends CLIP's zero-shot prediction capability to zero-shot segmentation. 
\citet{Han_2024_CVPR} leveraged features from DINOv2 and utilized LLMs as few-shot learners to achieve few-shot object detection.
Several works~\citep{qiu2024aligndiff, zhu2024unleashing, tan2023diffss} harness the generative capabilities of latent diffusion models for few-shot semantic segmentation.
In 3D, several studies~\citep{zhou2023uni3d, zhang2021pointclip} explored pretraining on large-scale point cloud datasets, achieving promising results in downstream tasks, e.g., one-shot part segmentation and point cloud classification.
While effective, these methods still lag significantly behind the successes demonstrated in image and language domains, due to limited 3D data.
We investigate one-shot learning with point clouds for object-to-object affordance grounding using VFMs, demonstrating geometric and semantic generalization.

\textbf{Affordance-Guided Robotic Manipulation.}
Affordance, as an intermediate representation that links perception and action, has been widely used for robotic manipulation.
In reinforcement learning, several works have explored using affordance as guidance for sample-efficient policy learning~\citep{borja2022affordance}, reward shaping~\citep{lee2024affordance, bahl2023affordances, geng2023rlafford}, and sub-goal specification~\citep{fang2023generalization, bahl2023affordances}.
In imitation learning, affordance is commonly used either as an additional conditional input~\citep{nasiriany2024rt, wu2024afforddp} or as an auxiliary learning objective~\citep{zha2021contrastively, mees23hulc2} to enhance generalization and improve sample efficiency.
In planning-based manipulation, recent efforts utilized affordance as a versatile representation for low-level planning.
RAM~\citep{kuang2024ram} adapt a retrieval-based approach for affordance transfer and action planning.
\citep{pmlr-v229-huang23b, huang2024rekep} used LLM-generated 3D affordance map or VLM-identified key points for low-level planning. RoboPoint~\citep{yuan2024robopoint} fine-tunes a VLM with affordance data to enable affordance reasoning capability for pick-and-place manipulation tasks.
Our work, on the other hand, formulates planning as an optimization problem while employing LLMs to automatically create constraint functions that use the learned affordance maps, which effectively link affordance with actions.
\vspace{-0.3cm}

\section{Problem Formulation}

We aim to address the problem of learning 3D object-to-object affordance under extreme data constraints, specifically, with only \textit{one training sample} for each affordance category. 
We formulate object-to-object (O2O) affordance grounding as predicting a functional map between two object point clouds in which an O2O affordance category is defined as being uniquely identified by an interaction verb, e.g., put or pour, not by a specific object pair.
Given a source object point cloud $P_s \in \mathbb{R}^{N_s \times (3+n)}$ and a target object point cloud $P_t \in \mathbb{R}^{N_t \times (3+n)}$, where $N_s$ and $N_t$ represent the number of points in each point cloud respectively, our goal is to predict affordance maps $A_s \in [0,1]^{N_s}$ and $A_t \in [0,1]^{N_t}$ that indicate the likelihood of O2O interaction at each point.
Each point is represented by its 3D coordinates $(x,y,z)$ and an $n$-dimensional semantic feature vector extracted from vision foundation models.
Our model $f_\theta$ with parameters $\theta$ maps a pair of input point clouds to their respective affordance maps:
$
f_\theta: (\mathbf{P}^{src}, \mathbf{P}^{tgt}) \mapsto (\mathbf{A}^{src}, \mathbf{A}^{tgt})
$

The training set consists of a set of $K$ interacting object pairs, each corresponding to a distinct affordance category: $\mathcal{D}_{\text{train}} = \{(\mathbf{P}^{src}_{i}, \mathbf{P}^{tgt}_{i}, \mathbf{A}^{src}_{i}, \mathbf{A}^{tgt}_{i})\}_{i=1}^{K}$ where each $(\mathbf{P}^{src}_{i}, \mathbf{P}^{tgt}_{i})$ represents a \textit{unique} object pair exhibiting the $i\text{-th}$ category of affordance, and each affordance category appears \textit{only once} in the training set.
At test time, the model is evaluated in a zero-shot manner on novel object pairs exhibiting the same set of affordance categories $\mathcal{D}_{\text{test}} = \{(\mathbf{P}^{src}_{j}, \mathbf{P}^{tgt}_{j})\}_{j=1}^{K}$, but with entirely unseen objects: $\mathcal{D}_{\text{train}} \cap \mathcal{D}_{\text{test}} = \varnothing$.
\vspace{-1em}
\section{Methodology}

\begin{figure}[t]
    \centering
    \begin{tabular}{cc}
        \includegraphics[width=0.95\textwidth]{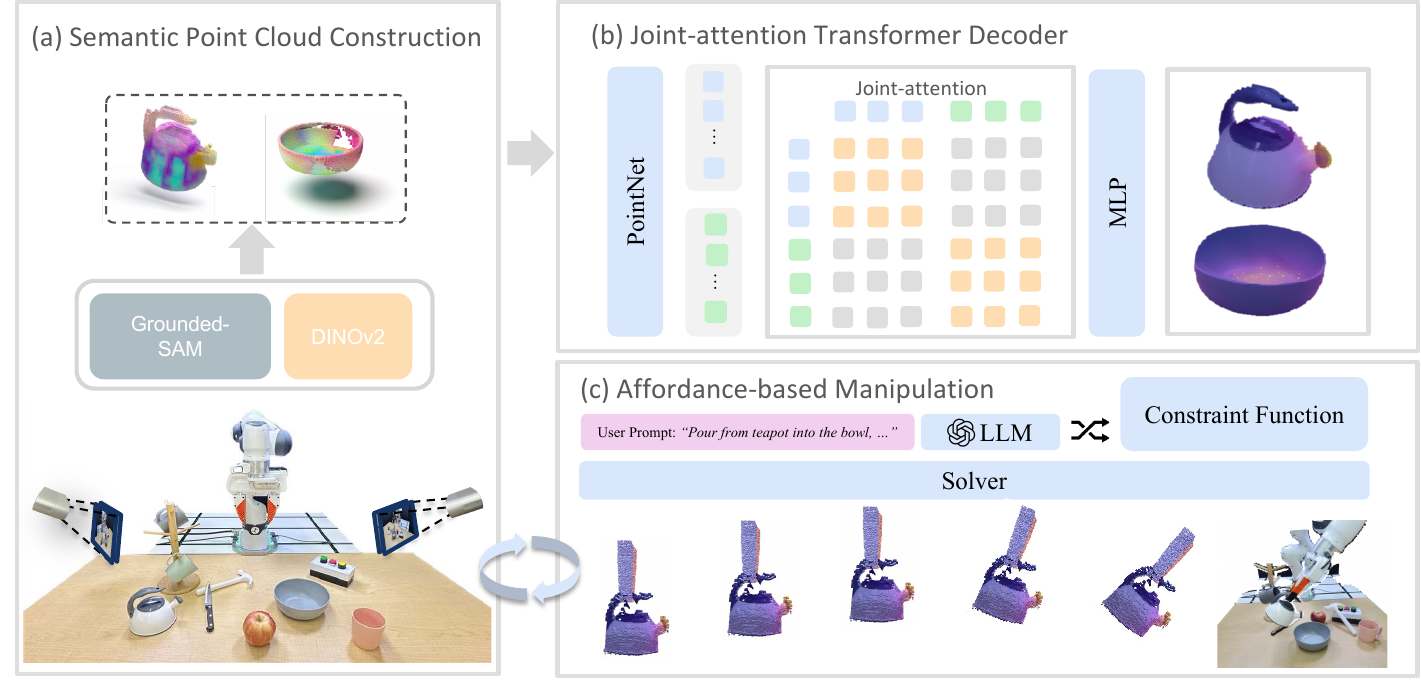}
    \end{tabular}
    \caption{\textbf{O$^3$Afford framework.} O$^3$Afford first (a) constructs semantic point clouds; (b) employs joint cross-attention between the source and target objects for decoding; lastly, (c) leverages LLMs for task-relevant constraint function generation and optimizes the target pose for robot execution.
    }
    \vspace{-0.5cm}
    \label{fig:method_pipeline}
\end{figure}

\subsection{Overview}
Our affordance grounding pipeline consists of three components as summarized in Fig.~\ref{fig:method_pipeline}.  
First, we construct 3D semantic features from DINOv2 for object point clouds.
Subsequently, our affordance grounding module takes these semantically-enriched point clouds as input and predicts the corresponding affordance maps.
To enable robotic manipulation, we leverage LLM for constraint function generation, which is integrated with optimization-based planning.

\subsection{Semantic Point Cloud}
To construct our 3D feature field for point cloud scenes, we draw inspiration from D$^3$Field~\citep{wang2024d3fields} for projecting 2D semantic features into 3D space using DINOv2~\citep{oquab2023dinov2}.
In their framework, multi-view RGBD images are processed to extract DINOv2 features, which are then projected onto arbitrary 3D coordinates by mapping them to each camera's image space, interpolating features, and fusing them across views.
Specifically, for a 3D point $\mathbf{x}$, we compute its projection $\mathbf{u}_i$ in the $i$-th camera view, determine the truncated depth difference $d_i = r_i - r_i'$ (where $r_i$ is sensor-captured depth and $r_i'$ is the interpolated depth), and assign visibility $v_i$ and weight $w_i$ to prioritize points near the surface.
These weights guide the fusion of semantic features $\mathbf{f}_i$ and instance masks $\mathbf{p}_i$ across $K$ views, yielding a unified 3D descriptor field.
We adapt this method by aligning multi-view RGB observations with the point cloud geometry, projecting DINOv2 features onto the 3D points, and fusing them to encode semantic information directly onto the point cloud representation.
This approach enables efficient and generalizable feature representation without additional training, supporting robust scene understanding in our work.
\subsection{One-Shot Affordance Grounding}
To enable effective affordance prediction from point clouds, we propose a joint-attention transformer decoder to predict affordance, which enables effective feature interaction across objects.
Our network processes paired point clouds representing a source object $\mathbf{P}^{src}\in\mathbb{R}^{B\times N\times 3}$ and a target object $\mathbf{P}^{tgt}\in\mathbb{R}^{B\times N\times 3}$, along with corresponding visual features extracted from DINOv2~\citep{oquab2023dinov2}, denoted by $\mathbf{F}^{\text{src}}, \mathbf{F}^{\text{tgt}}\in\mathbb{R}^{B\times N\times1024}$.
The output consists of per-point affordance scores optimized via binary cross-entropy (BCE) loss against ground-truth annotations during training.
Our model $f_{\theta}$ comprises two primary components: (1) a \textbf{Point Cloud Encoder}, (2) a \textbf{Joint-attention Transformer Decoder}.

\textbf{Point Cloud Encoder.}
The point cloud encoder first jointly processes concatenated point clouds and their DINOv2-derived features. 
Given the combined input $\mathbf{P}=[\mathbf{P}^{src}, \mathbf{P}^{tgt}]\in\mathbb{R}^{2B\times N\times 3}$ and features $\mathbf{F} = [\mathbf{F}^{\text{src}}, \mathbf{F}^{\text{tgt}}]\in\mathbb{R}^{2B\times N\times1024}$, we first tokenize the input using farthest point sampling (FPS) to select $T$ patch centers, then employ $k$-nearest neighbors for each patch center to construct the corresponding patches.
To effectively aggregate local geometric information into compact patch-level feature representations, we employ PointNet for feature extraction.
This process finally produces tokenized input $\mathbf{Z}\in\mathbb{R}^{2B\times T\times512}$. The object tokens are then concatenated with one-hot embeddings that distinguish between source object tokens and target object tokens for subsequent processing.

\textbf{Joint-attention Transformer Decoder.}
We apply a joint attention transformer to effectively capture geometric and semantic contextual dependencies across source and target objects. 
Specifically, we implement a cross-attention mechanism to enable dynamic feature interaction between objects.
Our multi-head attention module with 8 heads captures these contextual dependencies:
\[
\begin{aligned}
\mathbf{A}^{\text{src}} &= \text{CrossAttention}(\mathbf{Z}^{\text{src}}, \mathbf{Z}^{\text{tgt}}, \mathbf{Z}^{\text{tgt}}),
\mathbf{A}^{\text{tgt}} = \text{CrossAttention}(\mathbf{Z}^{\text{tgt}}, \mathbf{Z}^{\text{src}}, \mathbf{Z}^{\text{src}}).
\end{aligned}
\]
This bidirectional interaction module effectively encodes the complementary affordance relationships between interacting objects.

To generate per-point predictions, we interpolate the final patch-level embeddings back to individual points in the original point clouds using nearest-neighbor interpolation based on distance to patch centroids.
This process yields dense point-level embeddings $\mathbf{E}^{\text{src}}, \mathbf{E}^{\text{tgt}}\in\mathbb{R}^{B\times N\times512}$.
These dense embeddings are concatenated and processed through a lightweight MLP projection head to produce the final affordance map $\mathbf{A}\in[0,1]^{2B\times K \times N}$, representing predicted interaction probabilities, where the dimension $K$ corresponds to separate channels for each affordance type prediction.

\textbf{Training and Optimization.}
In training, we optimize our affordance grounding network using the binary cross-entropy (BCE) loss:
\[
\mathcal{L}_{\mathrm{BCE}} = -\frac{1}{N} \sum_{i=1}^{N} [y_i \cdot \log(\hat{y}_i) + (1 - y_i) \cdot \log(1 - \hat{y}_i)]
\]

During inference, the network directly outputs per-point affordance predictions, facilitating effective robotic manipulation planning and execution in real-world scenarios.

\subsection{Affordance-Based Manipulation}
\label{subsec:vlm_planning}
Given the predicted affordance on the objects' point clouds, which clearly reveals the spatial relationship between the two objects (e.g., how they should make contact), we formulate affordance-based manipulation as a constraint-based optimization problem.
Specifically, given a source object point cloud $\mathbf{P}^{src}$ and a target object point cloud $\mathbf{P}^{tgt}$, along with their respective affordance maps $\mathbf{A}^{src}$ and $\mathbf{A}^{tgt}$, we optimize a 6-DoF transformation $\mathbf{T} \in \mathrm{SE}(3)$ applied to the source object that aligns the objects for the intended task while satisfying several constraints formulated as:
\begin{equation}
\min_{\mathbf{T} \in \mathrm{SE}(3)} \quad \sum_{i=1}^{N} \lambda_i \cdot \mathcal{S}_i(\mathbf{P}^{src}, \mathbf{A}^{src}, \mathbf{P}^{tgt}, \mathbf{A}^{tgt}, \mathbf{T})
\end{equation}
\label{eqn:planning}
where $\mathcal{S}_i$ represents the $i$-th constraint function evaluating specific aspects of the task requirements (e.g., collision avoidance, etc.) and $\lambda_i$ is the weight reflecting the importance of each objective.

\textbf{Constraint Generation with LLMs.}
To make our system versatile and scalable, we utilize LLMs to generate task-specific constraint functions instead of manually designed rules.
Equipped with commonsense knowledge of daily tasks, LLMs can reason about how objects should interact and translate high-level semantics (e.g., ``pouring'', ``cutting'') into concrete geometric objectives.
Specifically, we prompt the LLM with a task description, and it outputs Python functions that take the object point clouds and their affordance maps as input.
These functions serve as score terms ${\mathcal{S}_i}$, capturing task-relevant spatial relationships such as alignment, contact, or insertion.
The resulting constraint scores are then combined in the framework of Eq.~\ref{eqn:planning} to optimize the final object pose $\mathbf{T} \in \mathrm{SE}(3)$ with off-the-shelf solvers.
We provide a detailed prompt template and example generated score functions in the Appendix~\ref{appendix:prompt_template} and \ref{appendix:example_contraint_function}.


\begin{figure}[t]
    \centering
    \includegraphics[width=0.9\textwidth]{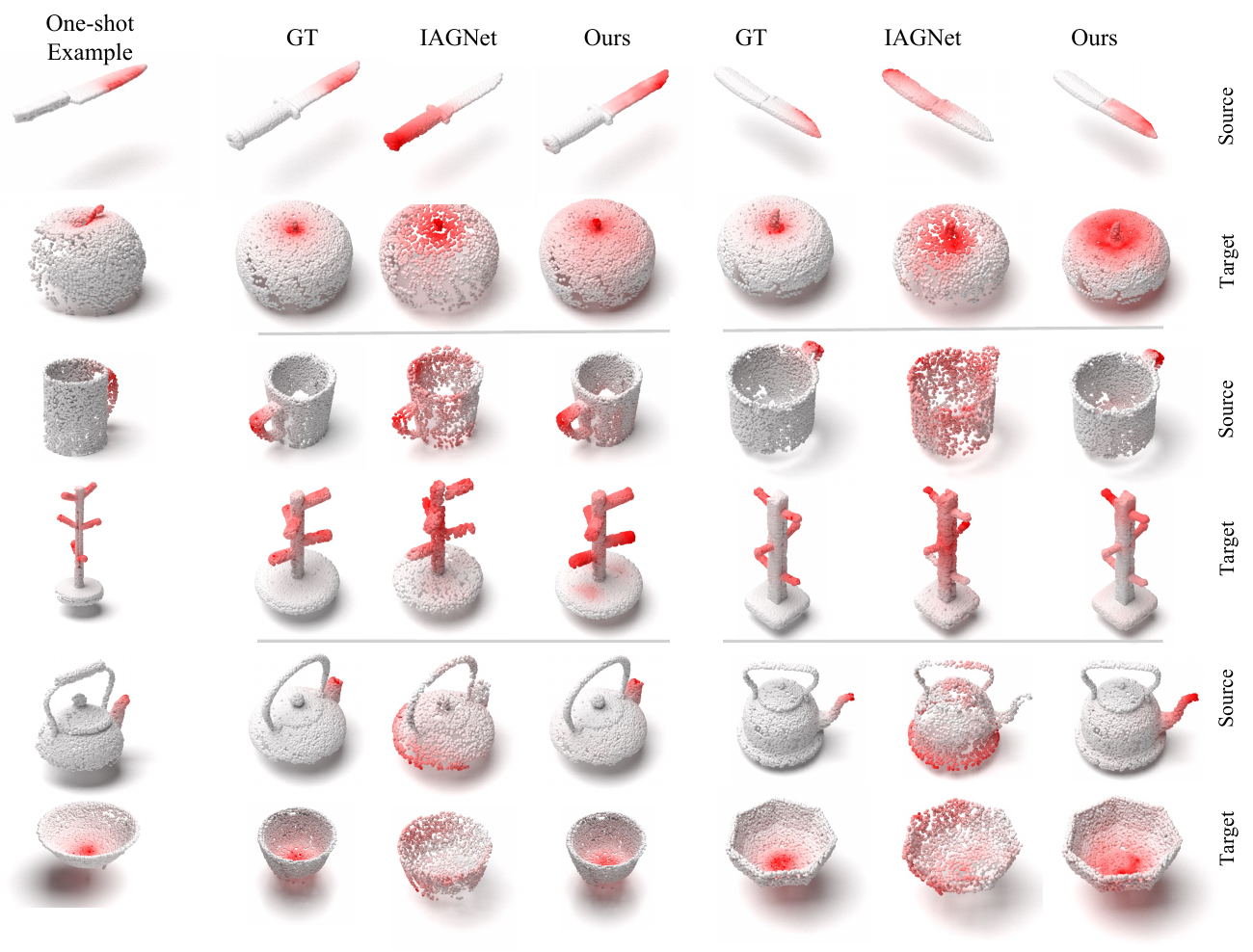}
    \caption{
    \textbf{Qualitative examples of the predicted object-to-object affordance in O$^3$Afford.}
    }
    \vspace{-0.5cm}
    \label{fig:affordance_qualitative}
\end{figure}

\section{Experiments}
We evaluate our method for both affordance prediction and robotic manipulation tasks.
We aim to answer three key research questions:
\textbf{(1)} How effectively does our method perform in object-to-object affordance grounding tasks?
\textbf{(2)} To what extent can our method generalize when training with only a single example for each affordance type?
\textbf{(3)} How effectively can our method improve downstream robotic manipulation?
We first demonstrate our experiment settings in Sec.~\ref{sec:exp_setup}.
We then address the above three questions through two stages: evaluating the accuracy and generalization capability of affordance grounding (Sec.~\ref{sec:affordance_grounding}) and validating our approach in both simulation and real-world robotic manipulation (Sec.~\ref{sec:manipulation}).

\subsection{Experiment Setup}
\label{sec:exp_setup}
Given the absence of high-quality object-to-object affordance grounding datasets, we annotate and construct our own dataset in simulation using SAPIEN~\citep{Xiang_2020_SAPIEN}.
The affordance map is generated in two steps: first, we have several user-assigned contact points on the point cloud, and then propagate the affordance label to other points based on distance following \citep{Deng_2021_CVPR}.

We conduct robot manipulation experiments in both simulated environments using SAPIEN~\citep{Xiang_2020_SAPIEN} and the real-world Franka Research 3.
In simulation, we position four stereo-depth sensors~\citep{liu2022activezero} from different viewpoints around the workspace.
In the real world, we position two Orbbec Femto Bolt cameras to get RGB-D observations.
We employ the LLM GPT-4o~\citep{achiam2023gpt} from OpenAI for constraint function generation in planning.
We design five tasks that require two objects to interact meaningfully with each other: \textit{pouring from teapot into bowl}, \textit{inserting toast into toaster}, \textit{pressing the button with hammer}, \textit{hanging mug onto mug tree} and \textit{cutting apple with knife} to evaluate our system's performance in varying contact geometries and spatial relationships between objects.
Our training dataset consists of a single pair of interacting objects for each affordance type.

\subsection{Affordance Grounding}
\label{sec:affordance_grounding}
\subsubsection{Main Results}
\label{sec:main_results}
\textbf{Baselines.}
We evaluate our affordance grounding module against 3 baselines: \textbf{O2O-Afford}~\citep{mo2022o2o}, a training-based 3D object-to-object method; \textbf{IAGNet~\citep{Yang_2023_ICCV}}, a method learning from 2D images; and \textbf{RoboPoint~\citep{yuan2024robopoint}}, a VLM-based approach.
We adopt four metrics during evaluation: \textbf{aIOU}~\citep{rahman2016optimizing}, \textbf{SIMilarity}~\citep{swain1991color}, \textbf{MAE}~\citep{willmott2005advantages}, and \textbf{AUC}~\citep{lobo2008auc}, which systematically measure spatial overlap, visual similarity, pixel-wise error, and overall prediction quality.

\begin{wrapfigure}{r}{0.5\textwidth}
    \vspace{-0.4cm}
    \centering
    \begin{threeparttable}
    \begin{tabular}{>{\centering\arraybackslash}p{1.8cm}|w{c}{0.8cm}w{c}{0.8cm}w{c}{0.8cm}w{c}{0.8cm}}
    \toprule
    \textbf{Method} & $\uparrow$ \textbf{IOU} & $\uparrow$ \textbf{SIM} & $\downarrow$ \textbf{MAE} & $\uparrow$ \textbf{AUC} \\
    \midrule
    O2O-Afford & 14.31 & 0.5123 & 0.1219 & 74.29 \\
    IAGNet & 16.89 & 0.5574 & 0.1402 & 73.30 \\
    RoboPoint & 11.84 & 0.4376 & 0.3344 & 59.78 \\
    \textbf{Ours} & \cellcolor{orange!20}\textbf{26.19} & \cellcolor{orange!20}\textbf{0.6387} & \cellcolor{orange!20}\textbf{0.0612} & \cellcolor{orange!20}\textbf{96.00} \\
    \bottomrule
    \end{tabular}
    \end{threeparttable}
    \captionof{table}{\textbf{Quantitative comparison on object-to-object affordance grounding methods.}}
    \vspace{-0.3cm}
    \label{tab:affordance_quantitative}
\end{wrapfigure}

\textbf{Results.}
Tab.~\ref{tab:affordance_quantitative} shows the quantitative comparison.
We also present our qualitative examples in Fig.~\ref{fig:affordance_qualitative}.
Each row in Fig.~\ref{fig:affordance_qualitative} represents a distinct affordance type, with three different affordance types illustrated.
For each affordance type, we present two corresponding rows: the upper showing the source object and the lower displaying the target object. We observed that RoboPoint, as a vision-language model, demonstrates capability in object localization but lacks the precision to infer fine-grained affordance regions on objects.
IAGNet exhibits the best performance among all baselines but suffers significantly from overfitting due to the one-shot training paradigm.
Our method significantly outperforms all baselines and demonstrates robust generalization.

\subsubsection{Generalization Analysis}
\label{sec:generalization_analysis}
Beyond the robust intra-class generalization under extreme data constraints demonstrated in Sec.~\ref{sec:main_results}, we also analyze the generalization capability of O$^3$Afford in this section.
We study two forms of generalization: \textit{generalization to different levels of occlusion} and \textit{cross-category generalization}, both of which are prevalent and essential in real-world manipulation scenarios.

\begin{wrapfigure}{r}{0.5\textwidth}
    \centering
    \vspace{-0.4cm}
    \includegraphics[width=0.5\textwidth]{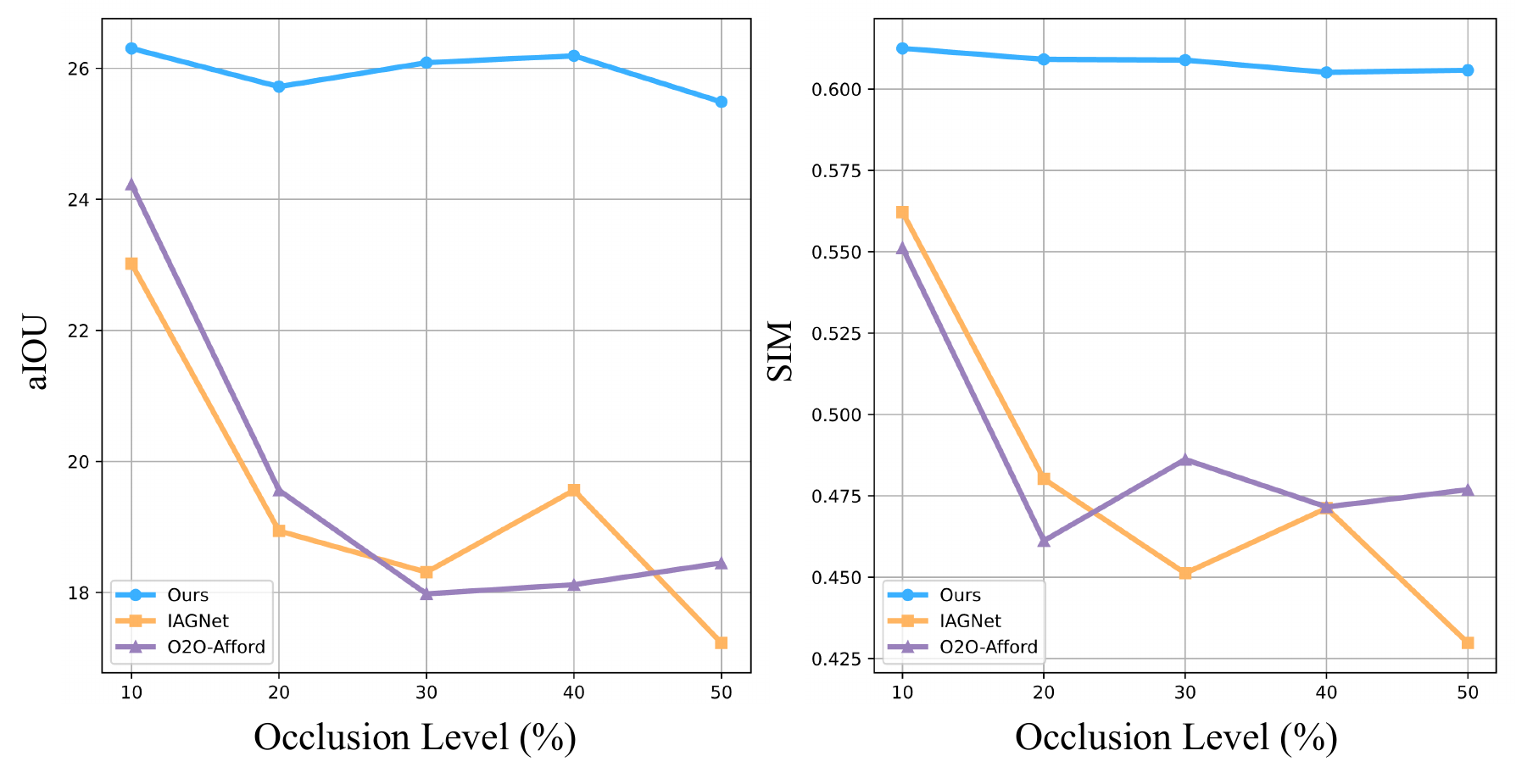}
   \caption{\textbf{Quantitative results under different occlusion levels}. Occlusion level (x-axis) ranges from 10\% to 50\% occluded point clouds.}
    \vspace{-0.5cm}
    \label{fig:occlusion_quantitative}
\end{wrapfigure}
\textbf{Occlusion Generalization.}
Rather than simulating occlusion by directly dropping points from the object point cloud, we evaluate generalization in a more realistic setting where objects are occluded by other objects (e.g., cubes).
We exclude RoboPoint in the comparison, as it is not a training-based method.
Fig.~\ref{fig:occlusion_quantitative} shows how aIOU and SIM change at different levels of occlusion.
Our method exhibits the least performance drop under varying levels of occlusion, whereas the two baselines suffer significant declines as occlusion increases due to larger geometric deviations from the training distribution.

\begin{figure}[t]
    \centering
    \includegraphics[width=1\textwidth]{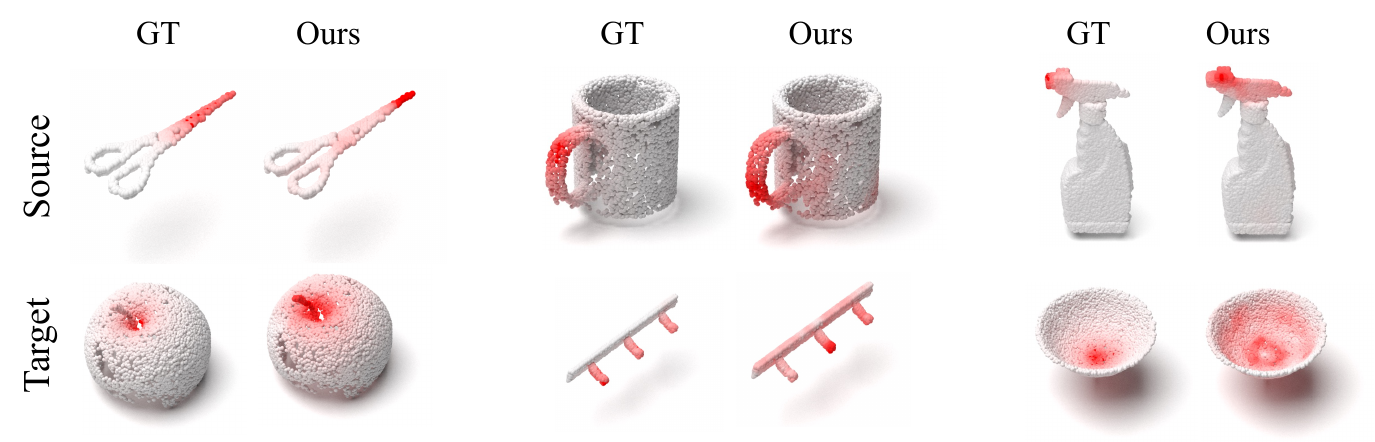}
    \caption{\textbf{Qualitative Results of Unseen Object Category.}}
    \vspace{-0.5cm}
    \label{fig:category_gen_qualitative}
\end{figure}

\textbf{Category-level Generalization.}
We consider object categories that are entirely unseen during training but may exhibit similar semantic or geometric functionality in manipulation tasks.
Fig.~\ref{fig:category_gen_qualitative} shows the example affordance predictions in a scissor for cutting, a coat rack for hanging, and a spray bottle for pouring.
These are unseen categories but exhibit similar semantic functionality.
Our model successfully predicts the affordance maps for these novel objects. This capability stems from both DINOv2's semantic features for identifying parts with similar semantics and the PointNet encoder for identifying similar local geometry, where we observe that latent patches with similar local geometries tend to be clustered together, as visualized in \appendref{appendix:latent_patch_embedding}.

\begin{table}[h]
    \centering
    \begin{threeparttable}
    \resizebox{\linewidth}{!}{
    \fontsize{12pt}{14pt}\selectfont 
    \begin{tabular}{
        p{1.5cm}|
        w{c}{1cm}w{c}{1cm}w{c}{1cm}|
        w{c}{1cm}w{c}{1cm}w{c}{1cm}|
        w{c}{1cm}w{c}{1cm}w{c}{1cm}|
        w{c}{1cm}w{c}{1cm}w{c}{1cm}|
        w{c}{1cm}w{c}{1cm}w{c}{1cm}
    }
    \toprule
    \textbf{Method} & 
    \multicolumn{3}{c|}{\textbf{Pour}} & 
    \multicolumn{3}{c|}{\textbf{Hang}} & 
    \multicolumn{3}{c|}{\textbf{Press}} & 
    \multicolumn{3}{c|}{\textbf{Insert}} & 
    \multicolumn{3}{c}{\textbf{Cut}} \\
    
    & 2-view & 3-view & 4-view & 
      2-view & 3-view & 4-view & 
      2-view & 3-view & 4-view & 
      2-view & 3-view & 4-view & 
      2-view & 3-view & 4-view \\
    \midrule
    Baseline & 0/10 & 2/10 & 2/10 & 0/10 & 1/10 & 0/10 & 4/10 & 5/10 & 5/10 & 1/10 & 2/10 & 5/10 & 3/10 & 5/10 & 4/10 \\
    Rekep    & 2/10 & 3/10 & 3/10    & 1/10 & 0/10 & 2/10    & 4/10 & 4/10 & 6/10    & 3/10 & 2/10 & 3/10    & 4/10 & 4/10 & 4/10   \\
    \textbf{Ours} & 
    \cellcolor{orange!20}\textbf{6/10} & \cellcolor{orange!20}\textbf{8/10} & \cellcolor{orange!20}\textbf{8/10} & 
    \cellcolor{orange!20}\textbf{3/10} & \cellcolor{orange!20}\textbf{3/10} & \cellcolor{orange!20}\textbf{5/10} & 
    \cellcolor{orange!20}\textbf{9/10} & \cellcolor{orange!20}\textbf{9/10} & \cellcolor{orange!20}\textbf{9/10} & 
    \cellcolor{orange!20}\textbf{5/10} & \cellcolor{orange!20}\textbf{7/10} & \cellcolor{orange!20}\textbf{8/10} & 
     \cellcolor{orange!20}\textbf{8/10} & \cellcolor{orange!20}\textbf{7/10} & \cellcolor{orange!20}\textbf{8/10} \\
    \bottomrule
    \end{tabular}
    }
    \end{threeparttable}
    \vspace{0.1cm}
    \caption{\textbf{Success rate of manipulation tasks in simulation.} Each method is evaluated under 2-view, 3-view, and 4-view setups over 10 trials.}
    \vspace{-0.7cm}
    \label{tab:sim_quantitative}
\end{table}

\subsection{Affordance-Guided Manipulation}
\label{sec:manipulation}

\textbf{Baselines.}
We evaluate our approach for robotic manipulation against two methods: (1) \textbf{\rekep{}}, which solves keypoint constraints through optimization for action planning; and (2) \textbf{Baseline, an ablated version} of our method that plans directly from object point clouds.
The evaluation is conducted across 5 tasks, with ten trials per task.
We record the success rate as the primary metric.

\textbf{Results in simulation.}
Tab.~\ref{tab:sim_quantitative} shows the success rate of each method.
We evaluate in settings with different numbers of camera views, ranging from four views, representing full observation, to as few as two views, representing partial observation with significant occlusion challenges.
Our method achieves the best overall performance, exhibiting the smallest performance drop when tested with fewer views, which highlights its robustness to occlusion.
In contrast, the baseline planning approach without affordance performs the worst due to its lack of awareness of object relationships during manipulation, underscoring the importance of incorporating affordance.
Compared to our method, \rekep{} exhibits two main drawbacks.
First, although \rekep{} can propose keypoints on objects for planning, it does not guarantee the task-awareness of the proposed keypoints, as evidenced by its lower performance compared to ours under the full observation setting.
Second, it is sensitive to occlusion, with its performance dropping significantly as the number of observation views decreases, making accurate keypoint proposal and tracking more challenging.

\begin{figure}[t]
    \centering
    \includegraphics[width=1\textwidth,height=5cm,trim=0pt 0pt 1pt 1pt,clip]{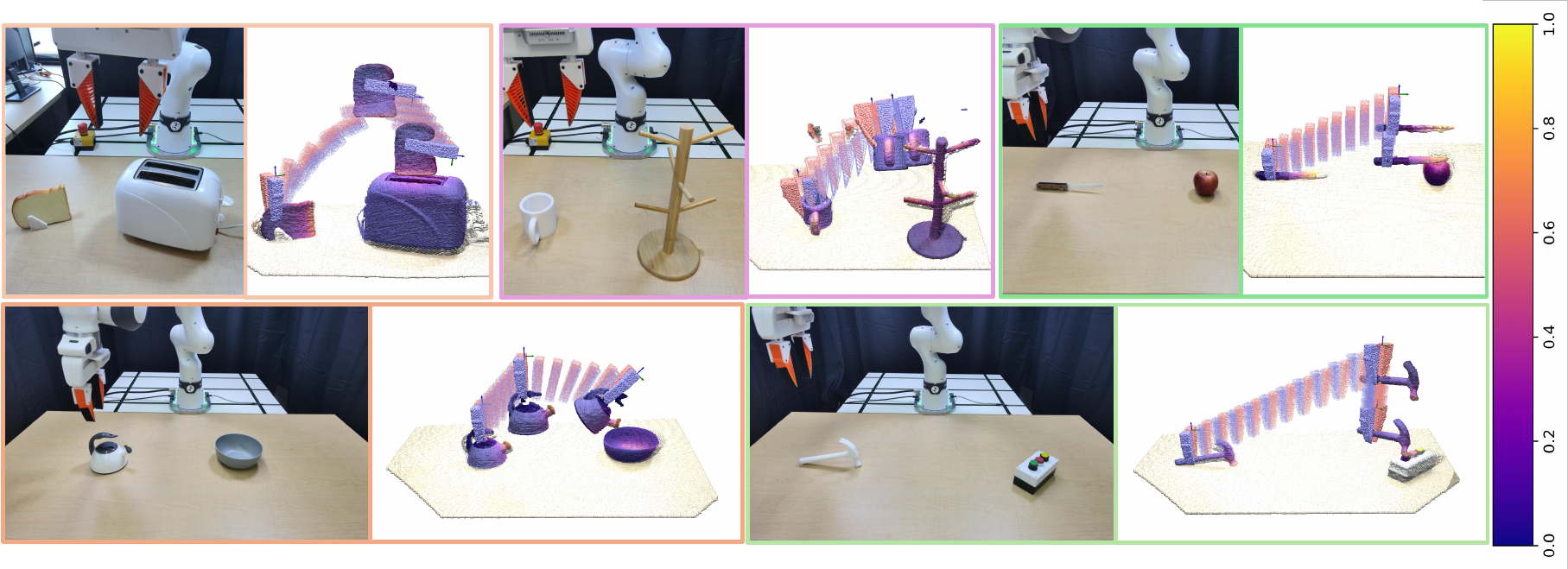}
    \caption{\textbf{Example execution of real-world manipulation tasks.} (Top) inserting, hanging, and cutting; (Bottom) pouring and pressing. The colors on the point clouds indicate the predicted affordance values as mapped to the colorbar on the right.}
    \label{fig:real_qualitative}
    \vspace{-0.5cm}
\end{figure}

\textbf{Results in real world.}
We evaluate our approach on 5 real-world manipulation tasks compared against the same baselines.
Tab.~\ref{tab:real_quantitative} shows their success rates.

\begin{wrapfigure}{r}{0.55\textwidth}
    \vspace{-0.4cm}
    \centering
    \begin{threeparttable}
    \small{
        \begin{tabular}{w{c}{1.1cm}|w{c}{0.8cm}w{c}{0.8cm}w{c}{0.8cm}w{c}{0.8cm}w{c}{0.8cm}}
        \toprule
        \textbf{Method} & \textbf{Pour} & \textbf{Hang} & \textbf{Press} & \textbf{Insert} & \textbf{Cut} \\
        \midrule
        Baseline & 2/10 & 0/10 & 2/10 & 3/10 & 3/10 \\
        \rekep{} & 3/10 & 2/10 & 5/10 & 4/10&  5/10 \\
        \textbf{Ours} & 
        \cellcolor{orange!20}\textbf{8/10} & 
        \cellcolor{orange!20}\textbf{5/10} & 
        \cellcolor{orange!20}\textbf{9/10} & 
        \cellcolor{orange!20}\textbf{8/10} & 
        \cellcolor{orange!20}\textbf{9/10} \\
        \bottomrule
        \end{tabular}
    }
    \end{threeparttable}
    \captionof{table}{\textbf{Success rate for different tasks on the real robot.} Each task is evaluated over 10 trials. Orange indicates improvements over the baseline.}
    \label{tab:real_quantitative}
    \vspace{-0.4cm}
\end{wrapfigure}

The result shows that affordance, as a mid-level representation, significantly enhances manipulation success rates.
In common tasks requiring two-object interaction, our method achieves approximately an 80\% success rate, while the baseline exhibits considerably poorer performance due to its inability to recognize functional properties from object point clouds.
In more complex and extended-horizon tasks such as hanging, the baseline fails in all trials, whereas our method maintains approximately 50\% success rate. 
We present qualitative examples in Fig.~\ref{fig:real_qualitative}.
The results demonstrate that affordance representation significantly enhances the feasibility of the generated constraint function for more reasonable interaction, as evidenced by the resulting coherent manipulation sequences.

\vspace{-0.25cm}
\section{Conclusion}
\vspace{-0.25cm}
We introduced O$^3$Afford, a one-shot learning framework that addresses object-to-object affordance grounding under extreme data constraints. By distilling semantic features from VFMs onto point clouds, our approach achieves effective intra-class and category-level generalization while maintaining robustness to occlusion. When integrated with LLMs to generate constraints for optimization-based robotic manipulation, our system demonstrates exceptional performance in complex real-world manipulation tasks, highlighting O$^3$Afford's potential for broader scenarios. Future work includes incorporating language instructions into our object-to-object affordance prediction pipeline to enable open-vocabulary affordance grounding.

\section*{Limitations}
While O$^3$Afford shows promise for figuring out where one 3D object can interact with another after seeing just one example, it still has some limitations. 
First, the system currently does not use language instructions to directly guide where it predicts the interaction should happen, which limits its ability to follow specific human directions. 
Second, the system's accuracy can decrease when parts of an object block the view of other parts (self-occlusion). This can result in an incomplete 3D model of the object and lead to incorrect predictions about interaction points on hidden parts, even if those hidden parts are important for the task. 
Finally, the practical performance of O$^3$Afford is constrained by the inherent limitations and noise present in current RGB-D sensing technologies, whether based on stereo vision, Time-of-Flight (ToF), structured light, or others. These sensors often produce imperfect depth data under various real-world conditions. For instance, stereo systems can struggle with textureless or reflective surfaces, while ToF sensors' accuracy can be affected by the target object's surface material (impacting light reflection) and potentially suffer from multipath interference in cluttered scenes. This underlying noise and potential inaccuracies in the input depth data can degrade the quality of the 3D point clouds used by our method, consequently limiting the precision and reliability of the final affordance predictions.

\section{Acknowledgments}
This research was partly supported by Delta Electronics Inc. and NSF CMMI-2443076.
We acknowledge Research Computing at the University of Virginia for providing the computational resources that made the results in this work possible.
We would like to thank Zezhou Cheng and Henry Kautz for their valuable feedback on the early draft of the paper.


\bibliography{main}

\begin{thebibliography}{60}
\providecommand{\natexlab}[1]{#1}
\providecommand{\url}[1]{\texttt{#1}}
\expandafter\ifx\csname urlstyle\endcsname\relax
  \providecommand{\doi}[1]{doi: #1}\else
  \providecommand{\doi}{doi: \begingroup \urlstyle{rm}\Url}\fi

\bibitem[Kaptelinin and Nardi(2012)]{kaptelinin2012affordances}
V.~Kaptelinin and B.~Nardi.
\newblock Affordances in hci: toward a mediated action perspective.
\newblock In \emph{Proceedings of the SIGCHI conference on human factors in computing systems}, pages 967--976, 2012.

\bibitem[Delitzas et~al.(2024)Delitzas, Takmaz, Tombari, Sumner, Pollefeys, and Engelmann]{delitzas2024scenefun3d}
A.~Delitzas, A.~Takmaz, F.~Tombari, R.~Sumner, M.~Pollefeys, and F.~Engelmann.
\newblock Scenefun3d: fine-grained functionality and affordance understanding in 3d scenes.
\newblock In \emph{Proceedings of the IEEE/CVF Conference on Computer Vision and Pattern Recognition}, pages 14531--14542, 2024.

\bibitem[Kuang et~al.(2024)Kuang, Ye, Geng, Mao, Deng, Guibas, Wang, and Wang]{kuang2024ram}
Y.~Kuang, J.~Ye, H.~Geng, J.~Mao, C.~Deng, L.~Guibas, H.~Wang, and Y.~Wang.
\newblock Ram: Retrieval-based affordance transfer for generalizable zero-shot robotic manipulation.
\newblock \emph{arXiv preprint arXiv:2407.04689}, 2024.

\bibitem[Li et~al.(2024)Li, Sun, Sevilla-Lara, and Jampani]{li2024one}
G.~Li, D.~Sun, L.~Sevilla-Lara, and V.~Jampani.
\newblock One-shot open affordance learning with foundation models.
\newblock In \emph{Proceedings of the IEEE/CVF Conference on Computer Vision and Pattern Recognition}, pages 3086--3096, 2024.

\bibitem[Luo et~al.(2022)Luo, Zhai, Zhang, Cao, and Tao]{luo2022learning}
H.~Luo, W.~Zhai, J.~Zhang, Y.~Cao, and D.~Tao.
\newblock Learning affordance grounding from exocentric images.
\newblock In \emph{Proceedings of the IEEE/CVF conference on computer vision and pattern recognition}, pages 2252--2261, 2022.

\bibitem[Luo et~al.(2023)Luo, Zhai, Zhang, Cao, and Tao]{luo2023leverage}
H.~Luo, W.~Zhai, J.~Zhang, Y.~Cao, and D.~Tao.
\newblock Leverage interactive affinity for affordance learning.
\newblock In \emph{Proceedings of the IEEE/CVF Conference on Computer Vision and Pattern Recognition}, pages 6809--6819, 2023.

\bibitem[Zhai et~al.(2022)Zhai, Luo, Zhang, Cao, and Tao]{zhai2022one}
W.~Zhai, H.~Luo, J.~Zhang, Y.~Cao, and D.~Tao.
\newblock One-shot object affordance detection in the wild.
\newblock \emph{International Journal of Computer Vision}, 130\penalty0 (10):\penalty0 2472--2500, 2022.

\bibitem[Zhao et~al.(2020)Zhao, Cao, and Kang]{zhao2020object}
X.~Zhao, Y.~Cao, and Y.~Kang.
\newblock Object affordance detection with relationship-aware network.
\newblock \emph{Neural Computing and Applications}, 32\penalty0 (18):\penalty0 14321--14333, 2020.

\bibitem[Mo et~al.(2022)Mo, Qin, Xiang, Su, and Guibas]{mo2022o2o}
K.~Mo, Y.~Qin, F.~Xiang, H.~Su, and L.~Guibas.
\newblock O2o-afford: Annotation-free large-scale object-object affordance learning.
\newblock In \emph{Conference on robot learning}, pages 1666--1677. PMLR, 2022.

\bibitem[Oquab et~al.(2023)Oquab, Darcet, Moutakanni, Vo, Szafraniec, Khalidov, Fernandez, Haziza, Massa, El-Nouby, et~al.]{oquab2023dinov2}
M.~Oquab, T.~Darcet, T.~Moutakanni, H.~Vo, M.~Szafraniec, V.~Khalidov, P.~Fernandez, D.~Haziza, F.~Massa, A.~El-Nouby, et~al.
\newblock Dinov2: Learning robust visual features without supervision.
\newblock \emph{arXiv preprint arXiv:2304.07193}, 2023.

\bibitem[Radford et~al.(2021)Radford, Kim, Hallacy, Ramesh, Goh, Agarwal, Sastry, Askell, Mishkin, Clark, Krueger, and Sutskever]{pmlr-v139-radford21a}
A.~Radford, J.~W. Kim, C.~Hallacy, A.~Ramesh, G.~Goh, S.~Agarwal, G.~Sastry, A.~Askell, P.~Mishkin, J.~Clark, G.~Krueger, and I.~Sutskever.
\newblock Learning transferable visual models from natural language supervision.
\newblock In M.~Meila and T.~Zhang, editors, \emph{Proceedings of the 38th International Conference on Machine Learning}, volume 139 of \emph{Proceedings of Machine Learning Research}, pages 8748--8763. PMLR, 18--24 Jul 2021.
\newblock URL \url{https://proceedings.mlr.press/v139/radford21a.html}.

\bibitem[Rombach et~al.(2022)Rombach, Blattmann, Lorenz, Esser, and Ommer]{rombach2022high}
R.~Rombach, A.~Blattmann, D.~Lorenz, P.~Esser, and B.~Ommer.
\newblock High-resolution image synthesis with latent diffusion models.
\newblock In \emph{Proceedings of the IEEE/CVF conference on computer vision and pattern recognition}, pages 10684--10695, 2022.

\bibitem[Radford et~al.(2021)Radford, Kim, Hallacy, Ramesh, Goh, Agarwal, Sastry, Askell, Mishkin, Clark, et~al.]{radford2021learning}
A.~Radford, J.~W. Kim, C.~Hallacy, A.~Ramesh, G.~Goh, S.~Agarwal, G.~Sastry, A.~Askell, P.~Mishkin, J.~Clark, et~al.
\newblock Learning transferable visual models from natural language supervision.
\newblock In \emph{International conference on machine learning}, pages 8748--8763. PmLR, 2021.

\bibitem[Wang et~al.(2024)Wang, Zhang, Li, Kelestemur, Driggs-Campbell, Wu, Fei-Fei, and Li]{wang2024d3fields}
Y.~Wang, M.~Zhang, Z.~Li, T.~Kelestemur, K.~Driggs-Campbell, J.~Wu, L.~Fei-Fei, and Y.~Li.
\newblock D$^3$fields: Dynamic 3d descriptor fields for zero-shot generalizable rearrangement.
\newblock In \emph{8th Annual Conference on Robot Learning}, 2024.

\bibitem[Zhou et~al.(2023)Zhou, Lei, Zhang, Liu, and Liu]{zhou2022zegclip}
Z.~Zhou, Y.~Lei, B.~Zhang, L.~Liu, and Y.~Liu.
\newblock Zegclip: Towards adapting clip for zero-shot semantic segmentation.
\newblock \emph{Proceedings of the IEEE/CVF Conference on Computer Vision and Pattern Recognition (CVPR)}, 2023.

\bibitem[Han and Lim(2024)]{Han_2024_CVPR}
G.~Han and S.-N. Lim.
\newblock Few-shot object detection with foundation models.
\newblock In \emph{Proceedings of the IEEE/CVF Conference on Computer Vision and Pattern Recognition (CVPR)}, pages 28608--28618, June 2024.

\bibitem[Zhang et~al.(2023)Zhang, Herrmann, Hur, Polania~Cabrera, Jampani, Sun, and Yang]{zhang2023tale}
J.~Zhang, C.~Herrmann, J.~Hur, L.~Polania~Cabrera, V.~Jampani, D.~Sun, and M.-H. Yang.
\newblock A tale of two features: Stable diffusion complements dino for zero-shot semantic correspondence.
\newblock \emph{Advances in Neural Information Processing Systems}, 36:\penalty0 45533--45547, 2023.

\bibitem[Myers et~al.(2015)Myers, Teo, Ferm{\"u}ller, and Aloimonos]{myers2015affordance}
A.~Myers, C.~L. Teo, C.~Ferm{\"u}ller, and Y.~Aloimonos.
\newblock Affordance detection of tool parts from geometric features.
\newblock In \emph{2015 IEEE international conference on robotics and automation (ICRA)}, pages 1374--1381. IEEE, 2015.

\bibitem[Sawatzky et~al.(2017)Sawatzky, Srikantha, and Gall]{sawatzky2017weakly}
J.~Sawatzky, A.~Srikantha, and J.~Gall.
\newblock Weakly supervised affordance detection.
\newblock In \emph{Proceedings of the IEEE Conference on Computer Vision and Pattern Recognition}, pages 2795--2804, 2017.

\bibitem[Jang et~al.(2024)Jang, Seo, and Chun]{jang2024intra}
J.~H. Jang, H.~Seo, and S.~Y. Chun.
\newblock Intra: Interaction relationship-aware weakly supervised affordance grounding.
\newblock In \emph{European Conference on Computer Vision}, pages 18--34. Springer, 2024.

\bibitem[Li et~al.(2023)Li, Jampani, Sun, and Sevilla-Lara]{li2023locate}
G.~Li, V.~Jampani, D.~Sun, and L.~Sevilla-Lara.
\newblock Locate: Localize and transfer object parts for weakly supervised affordance grounding.
\newblock In \emph{Proceedings of the IEEE/CVF Conference on Computer Vision and Pattern Recognition}, pages 10922--10931, 2023.

\bibitem[Qian et~al.(2024)Qian, Chen, Bai, Zhou, Tu, and Li]{qian2024affordancellm}
S.~Qian, W.~Chen, M.~Bai, X.~Zhou, Z.~Tu, and L.~E. Li.
\newblock Affordancellm: Grounding affordance from vision language models.
\newblock In \emph{Proceedings of the IEEE/CVF Conference on Computer Vision and Pattern Recognition}, pages 7587--7597, 2024.

\bibitem[Chen et~al.(2024)Chen, Cong, and Kan]{chen2024worldafford}
C.~Chen, Y.~Cong, and Z.~Kan.
\newblock Worldafford: Affordance grounding based on natural language instructions.
\newblock In \emph{2024 IEEE 36th International Conference on Tools with Artificial Intelligence (ICTAI)}, pages 822--828. IEEE, 2024.

\bibitem[Ju et~al.(2024)Ju, Hu, Zhang, Zhang, Jiang, and Xu]{ju2024robo}
Y.~Ju, K.~Hu, G.~Zhang, G.~Zhang, M.~Jiang, and H.~Xu.
\newblock Robo-abc: Affordance generalization beyond categories via semantic correspondence for robot manipulation.
\newblock In \emph{European Conference on Computer Vision}, pages 222--239. Springer, 2024.

\bibitem[Tong et~al.(2024)Tong, Opipari, Lewis, Zeng, and Jenkins]{tong2024oval}
E.~Tong, A.~Opipari, S.~Lewis, Z.~Zeng, and O.~C. Jenkins.
\newblock Oval-prompt: Open-vocabulary affordance localization for robot manipulation through llm affordance-grounding.
\newblock \emph{arXiv preprint arXiv:2404.11000}, 2024.

\bibitem[Deng et~al.(2021)Deng, Xu, Wu, Chen, and Jia]{Deng_2021_CVPR}
S.~Deng, X.~Xu, C.~Wu, K.~Chen, and K.~Jia.
\newblock 3d affordancenet: A benchmark for visual object affordance understanding.
\newblock In \emph{Proceedings of the IEEE/CVF Conference on Computer Vision and Pattern Recognition (CVPR)}, pages 1778--1787, June 2021.

\bibitem[Yang et~al.(2023)Yang, Zhai, Luo, Cao, Luo, and Zha]{Yang_2023_ICCV}
Y.~Yang, W.~Zhai, H.~Luo, Y.~Cao, J.~Luo, and Z.-J. Zha.
\newblock Grounding 3d object affordance from 2d interactions in images.
\newblock In \emph{Proceedings of the IEEE/CVF International Conference on Computer Vision (ICCV)}, pages 10905--10915, October 2023.

\bibitem[Yang et~al.(2024)Yang, Zhai, Luo, Cao, and Zha]{yang2024lemon}
Y.~Yang, W.~Zhai, H.~Luo, Y.~Cao, and Z.-J. Zha.
\newblock Lemon: Learning 3d human-object interaction relation from 2d images.
\newblock In \emph{Proceedings of the IEEE/CVF Conference on Computer Vision and Pattern Recognition}, pages 16284--16295, 2024.

\bibitem[Chu et~al.(2025)Chu, Deng, Lv, Chen, Li, Hao, and Nie]{chu20253d}
H.~Chu, X.~Deng, Q.~Lv, X.~Chen, Y.~Li, J.~Hao, and L.~Nie.
\newblock 3d-affordancellm: Harnessing large language models for open-vocabulary affordance detection in 3d worlds.
\newblock \emph{arXiv preprint arXiv:2502.20041}, 2025.

\bibitem[Li et~al.(2024)Li, Zhao, Xiao, Feng, Wang, and Chua]{Li_2024_CVPR}
Y.~Li, N.~Zhao, J.~Xiao, C.~Feng, X.~Wang, and T.-s. Chua.
\newblock Laso: Language-guided affordance segmentation on 3d object.
\newblock In \emph{Proceedings of the IEEE/CVF Conference on Computer Vision and Pattern Recognition (CVPR)}, pages 14251--14260, June 2024.

\bibitem[Brown et~al.(2020)Brown, Mann, Ryder, Subbiah, Kaplan, Dhariwal, Neelakantan, Shyam, Sastry, Askell, Agarwal, Herbert-Voss, Krueger, Henighan, Child, Ramesh, Ziegler, Wu, Winter, Hesse, Chen, Sigler, Litwin, Gray, Chess, Clark, Berner, McCandlish, Radford, Sutskever, and Amodei]{NEURIPS2020_1457c0d6}
T.~Brown, B.~Mann, N.~Ryder, M.~Subbiah, J.~D. Kaplan, P.~Dhariwal, A.~Neelakantan, P.~Shyam, G.~Sastry, A.~Askell, S.~Agarwal, A.~Herbert-Voss, G.~Krueger, T.~Henighan, R.~Child, A.~Ramesh, D.~Ziegler, J.~Wu, C.~Winter, C.~Hesse, M.~Chen, E.~Sigler, M.~Litwin, S.~Gray, B.~Chess, J.~Clark, C.~Berner, S.~McCandlish, A.~Radford, I.~Sutskever, and D.~Amodei.
\newblock Language models are few-shot learners.
\newblock In H.~Larochelle, M.~Ranzato, R.~Hadsell, M.~Balcan, and H.~Lin, editors, \emph{Advances in Neural Information Processing Systems}, volume~33, pages 1877--1901. Curran Associates, Inc., 2020.
\newblock URL \url{https://proceedings.neurips.cc/paper_files/paper/2020/file/1457c0d6bfcb4967418bfb8ac142f64a-Paper.pdf}.

\bibitem[Song et~al.(2022)Song, Dong, Zhang, Liu, and Wei]{song2022clip}
H.~Song, L.~Dong, W.-N. Zhang, T.~Liu, and F.~Wei.
\newblock Clip models are few-shot learners: Empirical studies on vqa and visual entailment.
\newblock \emph{arXiv preprint arXiv:2203.07190}, 2022.

\bibitem[Touvron et~al.(2023)Touvron, Martin, Stone, Albert, Almahairi, Babaei, Bashlykov, Batra, Bhargava, Bhosale, et~al.]{touvron2023llama}
H.~Touvron, L.~Martin, K.~Stone, P.~Albert, A.~Almahairi, Y.~Babaei, N.~Bashlykov, S.~Batra, P.~Bhargava, S.~Bhosale, et~al.
\newblock Llama 2: Open foundation and fine-tuned chat models.
\newblock \emph{arXiv preprint arXiv:2307.09288}, 2023.

\bibitem[Rombach et~al.(2021)Rombach, Blattmann, Lorenz, Esser, and Ommer]{rombach2021highresolution}
R.~Rombach, A.~Blattmann, D.~Lorenz, P.~Esser, and B.~Ommer.
\newblock High-resolution image synthesis with latent diffusion models, 2021.

\bibitem[Qiu et~al.(2024)Qiu, Wang, and Hauser]{qiu2024aligndiff}
R.-Z. Qiu, Y.-X. Wang, and K.~Hauser.
\newblock Aligndiff: aligning diffusion models for general few-shot segmentation.
\newblock In \emph{European Conference on Computer Vision}, pages 384--400. Springer, 2024.

\bibitem[Zhu et~al.(2024)Zhu, Liu, Luo, Jing, Chen, Xu, Wang, and Shen]{zhu2024unleashing}
M.~Zhu, Y.~Liu, Z.~Luo, C.~Jing, H.~Chen, G.~Xu, X.~Wang, and C.~Shen.
\newblock Unleashing the potential of the diffusion model in few-shot semantic segmentation.
\newblock \emph{arXiv preprint arXiv:2410.02369}, 2024.

\bibitem[Tan et~al.(2023)Tan, Chen, and Yan]{tan2023diffss}
W.~Tan, S.~Chen, and B.~Yan.
\newblock Diffss: Diffusion model for few-shot semantic segmentation.
\newblock \emph{arXiv preprint arXiv:2307.00773}, 2023.

\bibitem[Zhou et~al.(2024)Zhou, Wang, Ma, Liu, Huang, and Wang]{zhou2023uni3d}
J.~Zhou, J.~Wang, B.~Ma, Y.-S. Liu, T.~Huang, and X.~Wang.
\newblock Uni3d: Exploring unified 3d representation at scale.
\newblock In \emph{International Conference on Learning Representations (ICLR)}, 2024.

\bibitem[Zhang et~al.(2021)Zhang, Guo, Zhang, Li, Miao, Cui, Qiao, Gao, and Li]{zhang2021pointclip}
R.~Zhang, Z.~Guo, W.~Zhang, K.~Li, X.~Miao, B.~Cui, Y.~Qiao, P.~Gao, and H.~Li.
\newblock Pointclip: Point cloud understanding by clip.
\newblock \emph{arXiv preprint arXiv:2112.02413}, 2021.

\bibitem[Borja-Diaz et~al.(2022)Borja-Diaz, Mees, Kalweit, Hermann, Boedecker, and Burgard]{borja2022affordance}
J.~Borja-Diaz, O.~Mees, G.~Kalweit, L.~Hermann, J.~Boedecker, and W.~Burgard.
\newblock Affordance learning from play for sample-efficient policy learning.
\newblock In \emph{2022 International Conference on Robotics and Automation (ICRA)}, pages 6372--6378. IEEE, 2022.

\bibitem[Lee et~al.(2024)Lee, Xie, Fang, Pertsch, and Finn]{lee2024affordance}
O.~Y. Lee, A.~Xie, K.~Fang, K.~Pertsch, and C.~Finn.
\newblock Affordance-guided reinforcement learning via visual prompting.
\newblock \emph{arXiv preprint arXiv:2407.10341}, 2024.

\bibitem[Bahl et~al.(2023)Bahl, Mendonca, Chen, Jain, and Pathak]{bahl2023affordances}
S.~Bahl, R.~Mendonca, L.~Chen, U.~Jain, and D.~Pathak.
\newblock Affordances from human videos as a versatile representation for robotics.
\newblock In \emph{Proceedings of the IEEE/CVF Conference on Computer Vision and Pattern Recognition}, pages 13778--13790, 2023.

\bibitem[Geng et~al.(2023)Geng, An, Geng, Chen, Yang, and Dong]{geng2023rlafford}
Y.~Geng, B.~An, H.~Geng, Y.~Chen, Y.~Yang, and H.~Dong.
\newblock Rlafford: End-to-end affordance learning for robotic manipulation.
\newblock In \emph{2023 IEEE International Conference on Robotics and Automation (ICRA)}, pages 5880--5886. IEEE, 2023.

\bibitem[Fang et~al.(2023)Fang, Yin, Nair, Walke, Yan, and Levine]{fang2023generalization}
K.~Fang, P.~Yin, A.~Nair, H.~R. Walke, G.~Yan, and S.~Levine.
\newblock Generalization with lossy affordances: Leveraging broad offline data for learning visuomotor tasks.
\newblock In \emph{Conference on Robot Learning}, pages 106--117. PMLR, 2023.

\bibitem[Nasiriany et~al.(2024)Nasiriany, Kirmani, Ding, Smith, Zhu, Driess, Sadigh, and Xiao]{nasiriany2024rt}
S.~Nasiriany, S.~Kirmani, T.~Ding, L.~Smith, Y.~Zhu, D.~Driess, D.~Sadigh, and T.~Xiao.
\newblock Rt-affordance: Affordances are versatile intermediate representations for robot manipulation.
\newblock \emph{arXiv preprint arXiv:2411.02704}, 2024.

\bibitem[Wu et~al.(2024)Wu, Zhu, Huang, Zhu, Gu, Yu, Shi, and Wang]{wu2024afforddp}
S.~Wu, Y.~Zhu, Y.~Huang, K.~Zhu, J.~Gu, J.~Yu, Y.~Shi, and J.~Wang.
\newblock Afforddp: Generalizable diffusion policy with transferable affordance.
\newblock \emph{arXiv preprint arXiv:2412.03142}, 2024.

\bibitem[Zha et~al.(2021)Zha, Bhambri, and Guan]{zha2021contrastively}
Y.~Zha, S.~Bhambri, and L.~Guan.
\newblock Contrastively learning visual attention as affordance cues from demonstrations for robotic grasping.
\newblock In \emph{2021 IEEE/RSJ International Conference on Intelligent Robots and Systems (IROS)}, pages 7835--7842. IEEE, 2021.

\bibitem[Mees et~al.(2023)Mees, Borja-Diaz, and Burgard]{mees23hulc2}
O.~Mees, J.~Borja-Diaz, and W.~Burgard.
\newblock Grounding language with visual affordances over unstructured data.
\newblock In \emph{Proceedings of the IEEE International Conference on Robotics and Automation (ICRA)}, London, UK, 2023.

\bibitem[Huang et~al.(2023)Huang, Wang, Zhang, Li, Wu, and Fei-Fei]{pmlr-v229-huang23b}
W.~Huang, C.~Wang, R.~Zhang, Y.~Li, J.~Wu, and L.~Fei-Fei.
\newblock Voxposer: Composable 3d value maps for robotic manipulation with language models.
\newblock In J.~Tan, M.~Toussaint, and K.~Darvish, editors, \emph{Proceedings of The 7th Conference on Robot Learning}, volume 229 of \emph{Proceedings of Machine Learning Research}, pages 540--562. PMLR, 06--09 Nov 2023.
\newblock URL \url{https://proceedings.mlr.press/v229/huang23b.html}.

\bibitem[Huang et~al.(2024)Huang, Wang, Li, Zhang, and Fei-Fei]{huang2024rekep}
W.~Huang, C.~Wang, Y.~Li, R.~Zhang, and L.~Fei-Fei.
\newblock Rekep: Spatio-temporal reasoning of relational keypoint constraints for robotic manipulation.
\newblock \emph{arXiv preprint arXiv:2409.01652}, 2024.

\bibitem[Yuan et~al.(2024)Yuan, Duan, Blukis, Pumacay, Krishna, Murali, Mousavian, and Fox]{yuan2024robopoint}
W.~Yuan, J.~Duan, V.~Blukis, W.~Pumacay, R.~Krishna, A.~Murali, A.~Mousavian, and D.~Fox.
\newblock Robopoint: A vision-language model for spatial affordance prediction for robotics.
\newblock \emph{arXiv preprint arXiv:2406.10721}, 2024.

\bibitem[Xiang et~al.(2020)Xiang, Qin, Mo, Xia, Zhu, Liu, Liu, Jiang, Yuan, Wang, Yi, Chang, Guibas, and Su]{Xiang_2020_SAPIEN}
F.~Xiang, Y.~Qin, K.~Mo, Y.~Xia, H.~Zhu, F.~Liu, M.~Liu, H.~Jiang, Y.~Yuan, H.~Wang, L.~Yi, A.~X. Chang, L.~J. Guibas, and H.~Su.
\newblock {SAPIEN}: A simulated part-based interactive environment.
\newblock In \emph{The IEEE Conference on Computer Vision and Pattern Recognition (CVPR)}, June 2020.

\bibitem[Liu et~al.(2022)Liu, Yang, Tao, Chen, Zhang, Ran, Liu, and Su]{liu2022activezero}
I.~Liu, E.~Yang, J.~Tao, R.~Chen, X.~Zhang, Q.~Ran, Z.~Liu, and H.~Su.
\newblock Activezero: Mixed domain learning for active stereovision with zero annotation.
\newblock In \emph{Proceedings of the IEEE/CVF Conference on Computer Vision and Pattern Recognition}, pages 13033--13042, 2022.

\bibitem[Achiam et~al.(2023)Achiam, Adler, Agarwal, Ahmad, Akkaya, Aleman, Almeida, Altenschmidt, Altman, Anadkat, et~al.]{achiam2023gpt}
J.~Achiam, S.~Adler, S.~Agarwal, L.~Ahmad, I.~Akkaya, F.~L. Aleman, D.~Almeida, J.~Altenschmidt, S.~Altman, S.~Anadkat, et~al.
\newblock Gpt-4 technical report.
\newblock \emph{arXiv preprint arXiv:2303.08774}, 2023.

\bibitem[Rahman and Wang(2016)]{rahman2016optimizing}
M.~A. Rahman and Y.~Wang.
\newblock Optimizing intersection-over-union in deep neural networks for image segmentation.
\newblock In \emph{International symposium on visual computing}, pages 234--244. Springer, 2016.

\bibitem[Swain and Ballard(1991)]{swain1991color}
M.~J. Swain and D.~H. Ballard.
\newblock Color indexing.
\newblock \emph{International journal of computer vision}, 7\penalty0 (1):\penalty0 11--32, 1991.

\bibitem[Willmott and Matsuura(2005)]{willmott2005advantages}
C.~J. Willmott and K.~Matsuura.
\newblock Advantages of the mean absolute error (mae) over the root mean square error (rmse) in assessing average model performance.
\newblock \emph{Climate research}, 30\penalty0 (1):\penalty0 79--82, 2005.

\bibitem[Lobo et~al.(2008)Lobo, Jim{\'e}nez-Valverde, and Real]{lobo2008auc}
J.~M. Lobo, A.~Jim{\'e}nez-Valverde, and R.~Real.
\newblock Auc: a misleading measure of the performance of predictive distribution models.
\newblock \emph{Global ecology and Biogeography}, 17\penalty0 (2):\penalty0 145--151, 2008.

\bibitem[Tang et~al.(2025)Tang, Huang, Wang, Li, Yuan, Zhang, Wu, and Fei-Fei]{tang2025uad}
Y.~Tang, W.~Huang, Y.~Wang, C.~Li, R.~Yuan, R.~Zhang, J.~Wu, and L.~Fei-Fei.
\newblock Uad: Unsupervised affordance distillation for generalization in robotic manipulation.
\newblock \emph{arXiv preprint arXiv:2506.09284}, 2025.

\bibitem[Xiang et~al.(2024)Xiang, Lv, Xu, Deng, Wang, Zhang, Chen, Tong, and Yang]{xiang2024structured}
J.~Xiang, Z.~Lv, S.~Xu, Y.~Deng, R.~Wang, B.~Zhang, D.~Chen, X.~Tong, and J.~Yang.
\newblock Structured 3d latents for scalable and versatile 3d generation.
\newblock \emph{arXiv preprint arXiv:2412.01506}, 2024.

\end{thebibliography}
\clearpage
\appendix
\appendixpage

\section{Additional Results}
\label{appendix:additonal_results}

\subsection{Ablation}
We present ablation results on the model architecture to clarify the specific contributions of each module within our affordance model in ~\tabref{tab:ablation}. Specifically, we ablated three components: DINOv2 features (w/o DINOv2), PointNet encoder (w/o PointNet), and joint-attention (w/o Joint-Attn).

\begin{table}[htbp]
    \centering
    \small
    \begin{threeparttable}
    \begin{tabular}{>{\centering\arraybackslash}p{2.2cm}|w{c}{0.8cm}w{c}{0.8cm}w{c}{0.8cm}w{c}{0.8cm}}
    \toprule
    \textbf{Method} & $\uparrow$ \textbf{IOU} & $\uparrow$ \textbf{SIM} & $\downarrow$ \textbf{MAE} & $\uparrow$ \textbf{AUC} \\
    \midrule
    w/o DINOv2 & 6.59 & 0.5261 & 0.0384 & 69.16 \\
    w/o PointNet & 20.06 & 0.5923 & 0.0541 & 98.90 \\
    w/o Joint-Attn & 17.15 & 0.5717 & 0.0707 & 95.55 \\
    \textbf{Ours} & \cellcolor{orange!20}\textbf{26.19} & \cellcolor{orange!20}\textbf{0.6387} & \cellcolor{orange!20}\textbf{0.0612} & \cellcolor{orange!20}\textbf{96.00} \\
    \bottomrule
    \end{tabular}
    \end{threeparttable}
    \caption{\textbf{Model Ablations.}}
    \label{tab:ablation}
\end{table}

\subsection{Occlusion Qualitative Results}
\begin{figure}[h]
    \centering
    \includegraphics[width=1\textwidth, height=5cm]{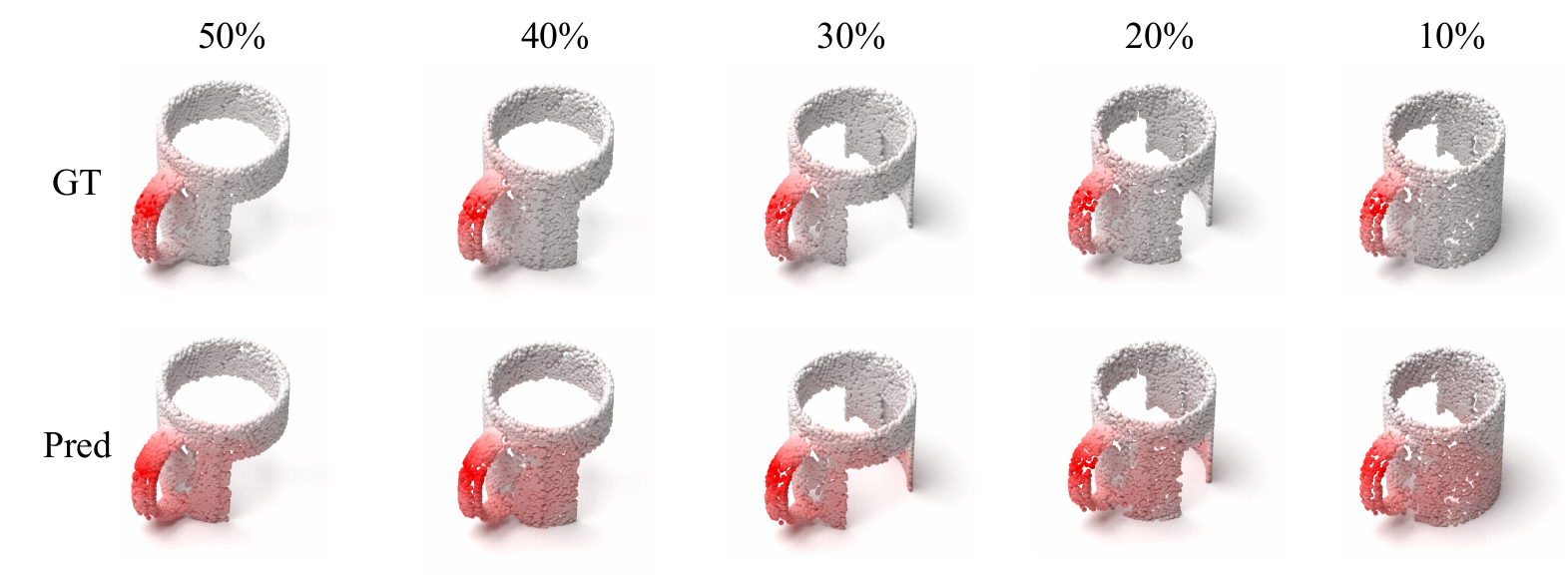}
    \caption{\textbf{Qualitative examples of different occlusion level.}}
    \label{fig:occlusion_qualitative}

\end{figure}

Fig.~\ref{fig:occlusion_qualitative} presents the qualitative examples of our occlusion experiment in the simulated environment. 
We show that our method, leveraging the generalizable semantic features of DINOv2, demonstrates robustness under varying levels of occlusion.
This includes extreme case with up to 50\% point cloud occluded which is considered the most difficult scenario in real-world applications.

\subsection{RoboPoint Qualitative Results}
\begin{figure}[h]
    \centering
    \includegraphics[width=1\textwidth]{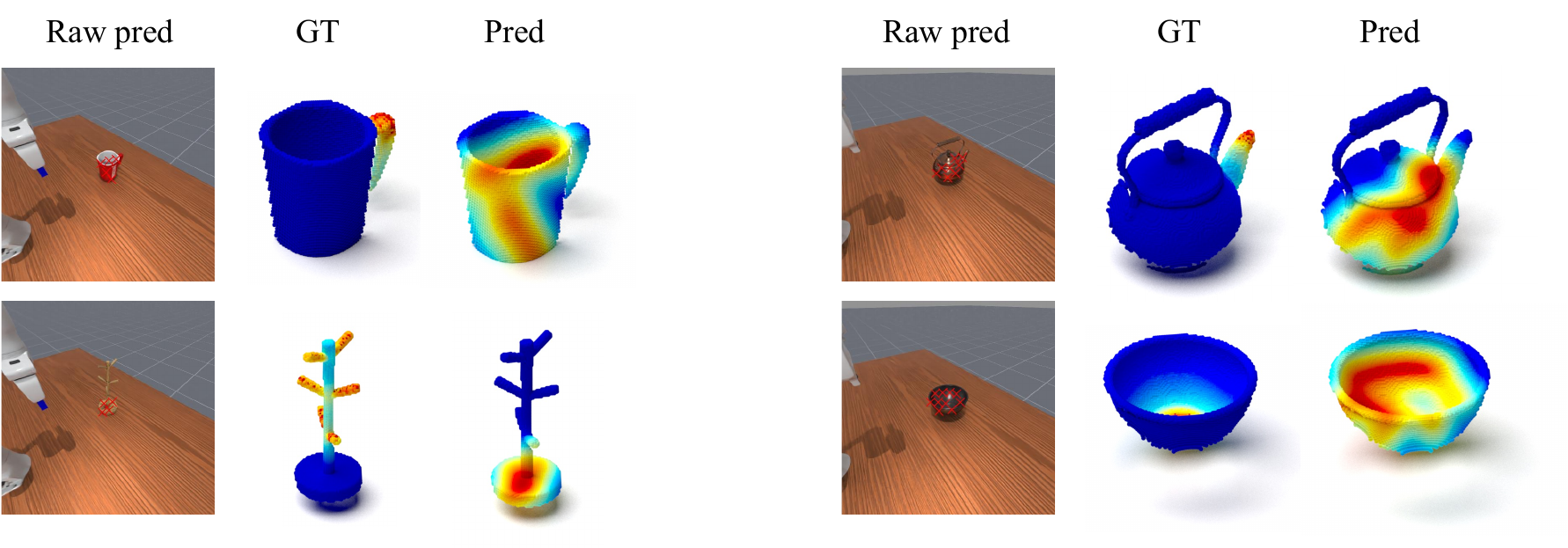}
    \caption{\textbf{Qualitative examples of RoboPoint predictions.}}
    \label{fig:robotpoint_qualitative}
    \vspace{-0.5cm}
\end{figure}

Fig.~\ref{fig:robotpoint_qualitative} shows the affordance prediction from RoboPoint~\citep{yuan2024robopoint}.
While RoboPoint. As a vision-language model, RoboPoint demonstrates strong object localization capabilities, but it lacks the precision required to infer fine-grained affordance regions on objects, as illustrated in Fig.~\ref{fig:robotpoint_qualitative}.
Most of its predictions successfully fall onto the area of the object, but don't onto the correct part of the object which hinders its application for more complex manipulation tasks.

\subsection{OOAL Qualitative Results}
\begin{figure}[h]
    \centering
    \includegraphics[width=0.9\textwidth]{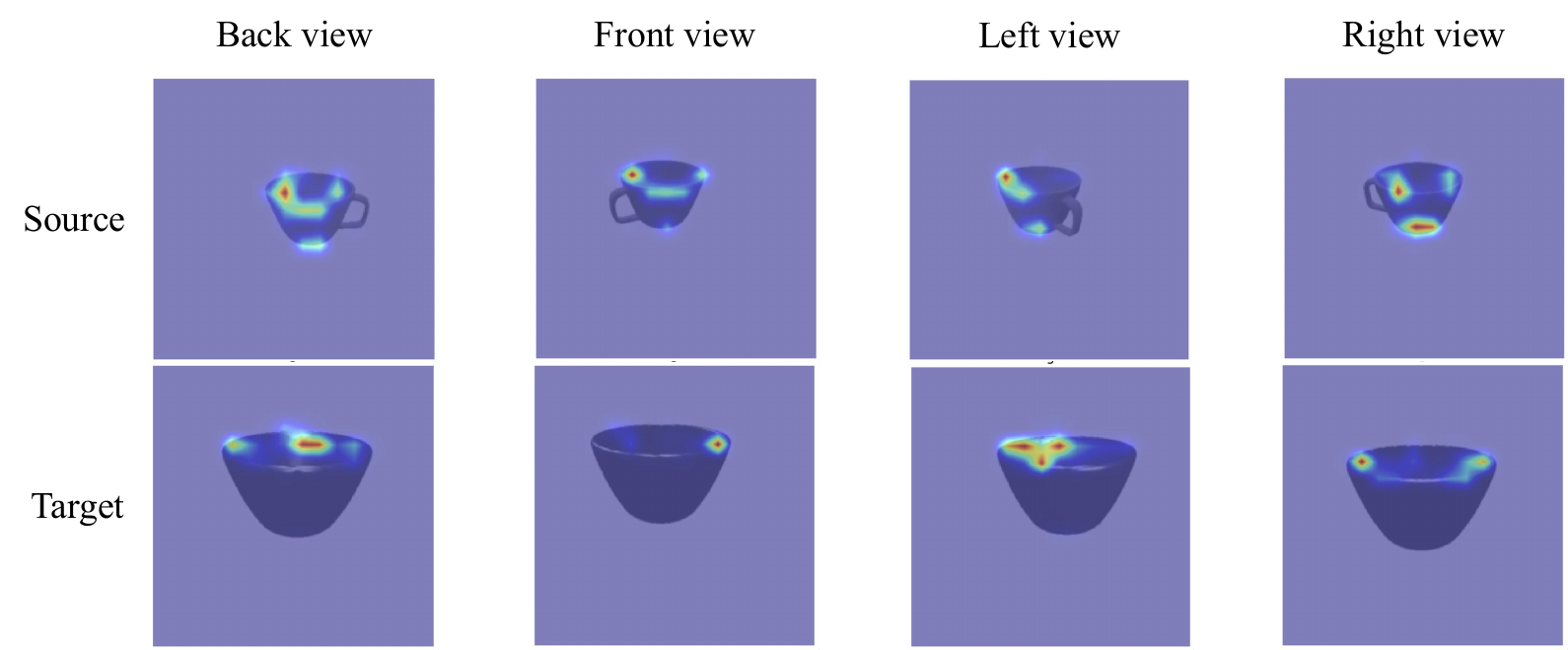}
    \caption{\textbf{Qualitative examples of OOAL predictions for four different views.} Each row renders the affordance heatmaps of the same source/target object across views.}
    \vspace{-2ex}
    \label{fig:ooal_qualitative}
\end{figure}

We highlight the limitations of recent work OOAL~\citep{li2024one}, which employs vision foundation models (VFMs) for 2D affordance grounding. OOAL exhibits two main drawbacks: (1) sensitivity to viewpoint changes, and (2) a noticeable sim-to-real performance gap.
We show the qualitative examples in Fig.~\ref{fig:ooal_qualitative}. We evaluate OOAL on data collected in the simulation environment using multiple viewpoints. The results reveal inconsistent predictions across viewpoints and a significant performance drop, despite the data is collected in the simulation. This highlights the benefit of affordance grounding in the 3D point cloud space, which reduces the sim-to-real gap and provides geometric awareness and robustness to viewpoint variations.

\subsection{Visualization of Latent Patch Embedding}
\label{appendix:latent_patch_embedding}
\begin{figure}[h]
    \centering
    \includegraphics[width=0.8\textwidth]{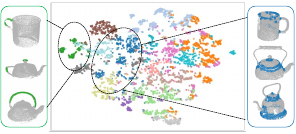}
    \caption{\textbf{t-SNE visualization of latent patch embeddings.}}
    \vspace{-0.5cm}
    \label{fig:t-sne visualization}
\end{figure}

For the PointNet-encoded patch embeddings, we observe that they encode the reusable geometric primitives across different objects as shown in our t-SNE plot (\figref{fig:t-sne visualization}), which shows that patches with similar local geometries tend to cluster together. We hypothesize that these patch embeddings contribute to the model's robustness to occlusion, as partial occlusion often leaves the local, functional parts unaffected.

\subsection{Manipulation Qualitative Results}
\begin{figure}[h]
    \centering
    \includegraphics[width=1\textwidth]{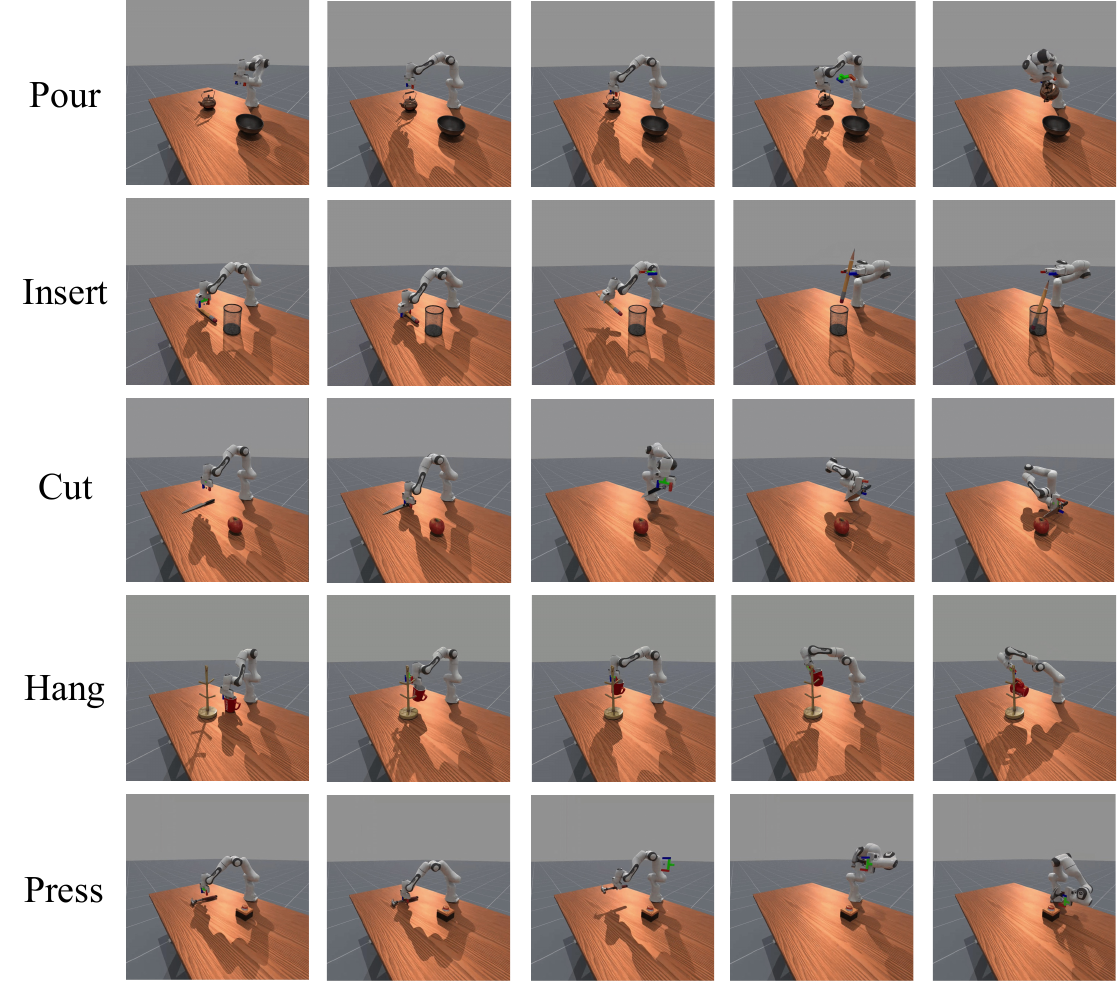}
    \caption{\textbf{Qualitative results in simulation environment.}}
    \vspace{-0.5cm}
    \label{fig:sim_manipulation_qualitative}
\end{figure}

Fig.~\ref{fig:sim_manipulation_qualitative} shows the example executions of the different types of actions in the simulator using our proposed method.

\clearpage
\section{Implementation Details}
\label{appendix:implementation_details}
\subsection{Baselines}
\paragraph{RoboPoint}
RoboPoint~\citep{yuan2024robopoint} is a vision-language model finetuned on an image-language affordance dataset. To apply it to a point cloud, we first prompt RoboPoint to predict affordance pixels in the image, which are then projected onto the point cloud using the corresponding depth information. Across all views, we aggregate these predictions by averaging the affordance scores assigned from different views weighted by depth consistency across views.
\paragraph{IAGNet} IAGNet~\citep{Yang_2023_ICCV} is a training-based method which takes an image, with an annotated object bounding box, and an object point cloud as input. To adapt the model for the object-to-object setting, we use the same interaction image, with bounding boxes annotated for both the source and target objects during training and testing, for each interacting object pair. This modification preserves the core idea of extracting semantic interaction information from 2D images without altering the model architecture.

\paragraph{UAD} UAD~\citep{tang2025uad} is a concurrent work that also projects features from DINOv2 onto the point cloud to segment out the parts and uses vision-language models for affordance auto-annotation. Our work diverges from UAD in task settings.

\subsection{Model Details}
\paragraph{Point Cloud Encoder}
We employ a patch-based architecture for point cloud encoding that processes the input through local grouping and feature extraction. The encoder first groups points using kNN, then processes each local patch through a specialized patch encoder, and finally incorporates positional information through learnable embeddings. This design enables effective capture of both local geometric structures and global spatial relationships.

\begin{table}[h]
\centering
\begin{tabular}{ll}
\hline
Hyperparameter & Value \\
\hline
Output dimension & 512 \\
Number of groups & 256 \\
Group size & 64 \\
Group radius & 0.15 \\
Position embedding dimension & 512 \\
Patch encoder hidden dims & [784, 512] \\
\hline
\end{tabular}
\vspace{0.2cm}
\caption{Hyperparameters of the point cloud encoder.}
\vspace{-0.5cm}
\end{table}

\paragraph{Decoder}
We employ a transformer-based decoder which additionally adopts joint-attention to enable feature communication between the source and target objects. We provide detailed hyperparameters in Table~\ref{tab:decoder_details}.

\begin{table}[h]
\centering
\begin{tabular}{ll}
\hline
Hyperparameter & Value \\
\hline
Number of attention heads & 8 \\
Attention head dimension & 512 \\
Dropout & 0.1 \\
Cross attention dimension & 512 \\
Point cloud embedding dimension & 512 \\
Activation function & GELU \\
Output MLP dimensions & [512, 256] \\
Normalization type & LayerNorm \\
Normalization epsilon & 1e-5 \\
\hline
\end{tabular}
\vspace{0.2cm}
\caption{Hyperparameters of the decoder.}
\label{tab:decoder_details}
\end{table}

\subsection{Prompt Template}
\label{appendix:prompt_template}
We use GPT-4o for constraint function generation. Note that our pipeline can be integrated with any advanced LLMs for constraint function generation. We provide an example prompt template below. 

\begin{tcolorbox}[title={Prompt Template for Certain  Constraint}]
You are given two 3D objects represented as point clouds. Each point is associated with an affordance score predicted by a perception model. Your task is to propose a constraint function for the specified affordance type that evaluates how well the source object can interact with the target object.

\small
\textbf{Inputs:}
\begin{itemize}
  \item \texttt{Source Object Name}: [SRC\_OBJECT\_NAME]
  \item \texttt{Target Object Name}: [TGT\_OBJECT\_NAME]
  \item \texttt{Interaction Type}: [AFFORDANCE] (e.g., pour, hang, press, cut, put, plug in)
  \item \texttt{Source Point Cloud}: $\{(x_i, y_i, z_i)\}_{i=1}^N$, with affordance scores $\{a_i\}_{i=1}^N$
  \item \texttt{Target Point Cloud}: $\{(x_j, y_j, z_j)\}_{j=1}^M$, with affordance scores $\{b_j\}_{j=1}^M$
\end{itemize}

\textbf{Task:}
Generate a constraint function that evaluates the quality of an affordance-specific interaction between source and target objects. The function should consider high-affordance regions, interaction-specific spatial constraints, and physical plausibility.

\textbf{Code Skeleton:}
\begin{verbatim}
def compute_alignment_score(src_aff, tgt_aff, src_pcd, tgt_pcd):
    score = 0
    """
    # TODO: Implement affordnace alignment contraint
    
    return score
\end{verbatim}

\textbf{Constraints:}
\begin{itemize}
  \item The function should use the information from high-affordance regions of both objects
  \item The evaluation must reflect the semantic meaning of the specified affordance type
  \item Consideration should be given to the physical feasibility of the interaction
  \item All constraints should be combined into a single cost value (lower is better)
\end{itemize}

\end{tcolorbox}

\subsection{Example Constraint Functions}
\label{appendix:example_contraint_function}
\begin{figure}[h]
    \centering
    \includegraphics[width=1\textwidth]{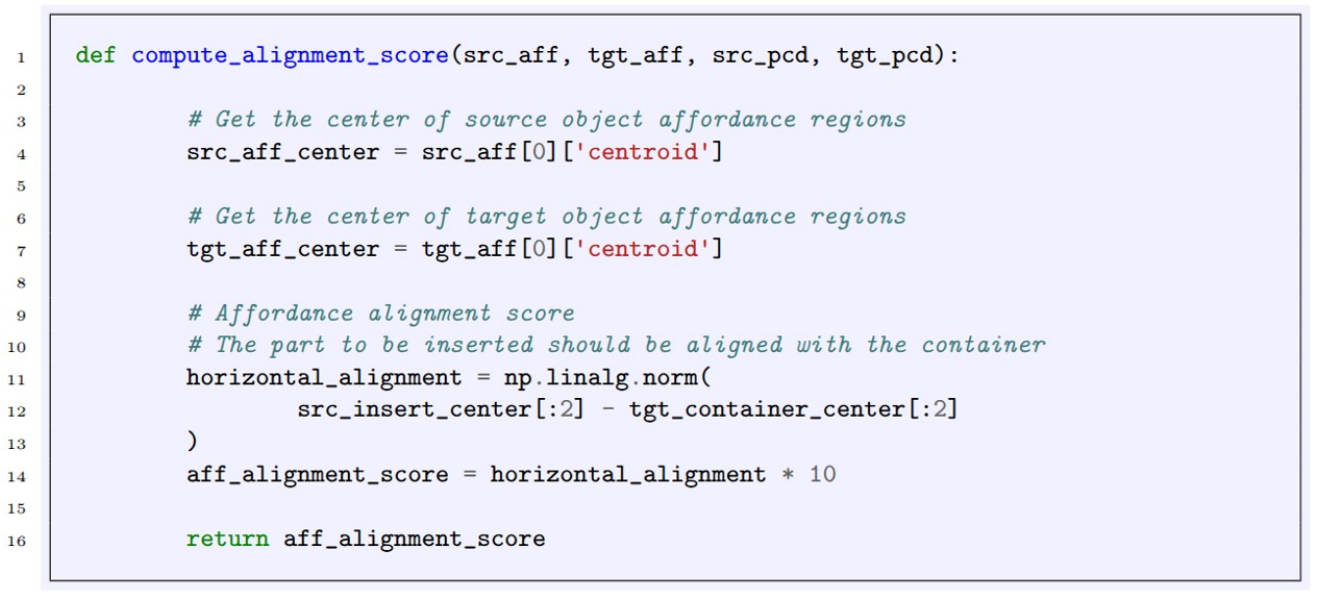}
    \caption{Example of the generated function for object alignment.}
    \vspace{-0.5cm}
    \label{fig:example_contraint_function}
\end{figure}

We provide an example of a generated constraint function in Fig.~\ref{fig:example_contraint_function}. A list of LLM-generated constraint functions with its corresponding weights is listed in~\tabref{tab:constraint_functions_list}.

\begin{table}[htbp]
  \centering
  
  \small
  
  \rowcolors{2}{gray!15}{white} 
  
  \begin{tabular}{@{} p{1.5cm} p{4.5cm} p{4.5cm} p{1.5cm} @{}}
    \toprule
    \textbf{Task} & \textbf{Primary Constraints} & \textbf{Evaluation Components} & \textbf{Weights} \\
    \midrule
    
    \textbf{Pour}   
    & \begin{itemize}[leftmargin=*, topsep=0pt, partopsep=0pt, itemsep=0pt, parsep=0pt]
        \item Position (above)
        \item Orientation (tilted 30-60°)
        \item Distance (medium clearance)
      \end{itemize}
    & \begin{itemize}[leftmargin=*, topsep=0pt, partopsep=0pt, itemsep=0pt, parsep=0pt]
        \item Affordance alignment
        \item Position constraint
        \item Orientation constraint
        \item Clearance
      \end{itemize}
    & \begin{itemize}[leftmargin=*, topsep=0pt, partopsep=0pt, itemsep=0pt, parsep=0pt]
        \item 0.3
        \item 0.2
        \item 0.3
        \item 0.1
      \end{itemize} \\
      
    \textbf{Hang}
    & \begin{itemize}[leftmargin=*, topsep=0pt, partopsep=0pt, itemsep=0pt, parsep=0pt]
        \item Position (contact)
        \item Orientation (hooked)
        \item Stability (CoM below support)
      \end{itemize}
    & \begin{itemize}[leftmargin=*, topsep=0pt, partopsep=0pt, itemsep=0pt, parsep=0pt]
        \item Affordance alignment
        \item Contact quality
        \item Stability
        \item Collision
      \end{itemize}
    & \begin{itemize}[leftmargin=*, topsep=0pt, partopsep=0pt, itemsep=0pt, parsep=0pt]
        \item 0.3
        \item 0.3
        \item 0.3
        \item 0.1
      \end{itemize} \\
      
    \textbf{Cut}
    & \begin{itemize}[leftmargin=*, topsep=0pt, partopsep=0pt, itemsep=0pt, parsep=0pt]
        \item Position (aligned)
        \item Orientation (perpendicular)
        \item Distance (contact)
      \end{itemize}
    & \begin{itemize}[leftmargin=*, topsep=0pt, partopsep=0pt, itemsep=0pt, parsep=0pt]
        \item Affordance alignment
        \item Position constraint
        \item Collision
      \end{itemize}
    & \begin{itemize}[leftmargin=*, topsep=0pt, partopsep=0pt, itemsep=0pt, parsep=0pt]
        \item 0.4
        \item 0.4
        \item 0.2
      \end{itemize} \\
      
    \textbf{Press}
    & \begin{itemize}[leftmargin=*, topsep=0pt, partopsep=0pt, itemsep=0pt, parsep=0pt]
        \item Position (aligned)
        \item Orientation (perpendicular)
        \item Distance (contact)
      \end{itemize}
    & \begin{itemize}[leftmargin=*, topsep=0pt, partopsep=0pt, itemsep=0pt, parsep=0pt]
        \item Affordance alignment
        \item Position constraint
        \item Orientation constraint
        \item Collision
      \end{itemize}
    & \begin{itemize}[leftmargin=*, topsep=0pt, partopsep=0pt, itemsep=0pt, parsep=0pt]
        \item 0.4
        \item 0.3
        \item 0.2
        \item 0.1
      \end{itemize} \\
      
    \textbf{Insert}
    & \begin{itemize}[leftmargin=*, topsep=0pt, partopsep=0pt, itemsep=0pt, parsep=0pt]
        \item Position (inside)
        \item Orientation (aligned)
        \item Distance (inside container)
      \end{itemize}
    & \begin{itemize}[leftmargin=*, topsep=0pt, partopsep=0pt, itemsep=0pt, parsep=0pt]
        \item Affordance alignment
        \item Position constraint
        \item Orientation constraint
        \item Collision
      \end{itemize}
    & \begin{itemize}[leftmargin=*, topsep=0pt, partopsep=0pt, itemsep=0pt, parsep=0pt]
        \item 0.3
        \item 0.4
        \item 0.2
        \item 0.1
      \end{itemize} \\
      
    \bottomrule
  \end{tabular}
  \vspace{0.2cm}
  \caption{Generated Constraint Functions for Different Manipulation Tasks}
  \label{tab:constraint_functions_list}
\end{table}

\section{Failure Cases Analysis}
\begin{figure}[h]
    \centering
    \includegraphics[width=1\textwidth, height=3cm]{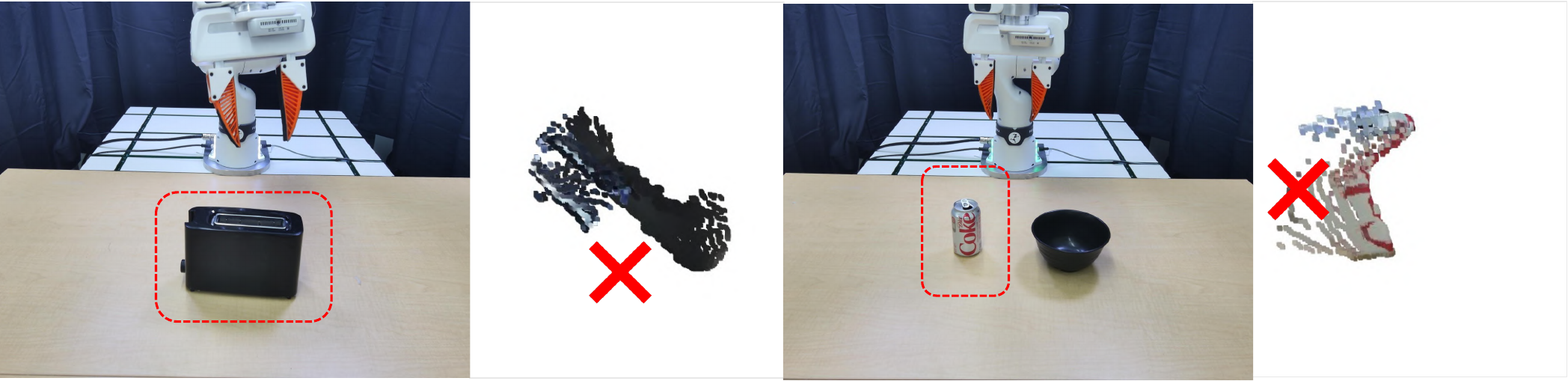}
    \caption{Example of incorrect point cloud.}
    \vspace{-0.5cm}
    \label{fig:example_sensor_noise}
\end{figure}

We present an empirical analysis of failure cases in this section. Our manipulation pipeline involves three key stages: constructing a semantic point cloud from multiview RGB-D observations, grounding affordance on the partially reconstructed point cloud, and executing actions via a low-level motion planner. As a result, failure cases primarily fall into three categories: (1) inaccurate point cloud reconstruction due to sensor noise, (2) incorrect affordance predictions caused by object self-occlusion and unobservable regions, and (3) inverse kinematics (IK) failure during motion planning.
\paragraph{Sensor noise}
We use the Orbbec Femto Bolt camera, a Time-of-Flight-based sensor, for capturing RGB-D observations.
It provides significantly higher-quality depth compared to stereo-based sensors.
However, it may still produce erroneous depth measurements when object materials are highly reflective or when objects are too dark in color, preventing the sensor from receiving sufficient reflected light to compute accurate depth. We provide an example of an incorrectly reconstructed point cloud in Fig.~\ref{fig:example_sensor_noise}.

\paragraph{Object Self-occlusion}
Our pipeline assumes that affordance regions, such as the mouth of a teapot, are not occluded. However, in practice, these regions can be self-occluded by the object itself.
For example, the bottom of a bottle is difficult to observe even with a top-down camera view. 
This limitation could potentially be addressed by recent advances in single-view or sparse-view reconstruction methods (e.g., TRELLIS~\citep{xiang2024structured}), which are orthogonal to our approach. We leave the integration of such techniques for future work.

\paragraph{IK Failure}
Since we directly optimize the final interaction pose of the object, the resulting pose may be infeasible for execution due to IK failure in the motion planner. We use a screw-based planner, and most IK failures occur when the final pose requires drastic rotational movements. We provide some visualizations of IK failure cases in Fig.~\ref{fig:example_ik_failure}.
\begin{figure}[h]
    \centering
    \includegraphics[width=0.9\textwidth]{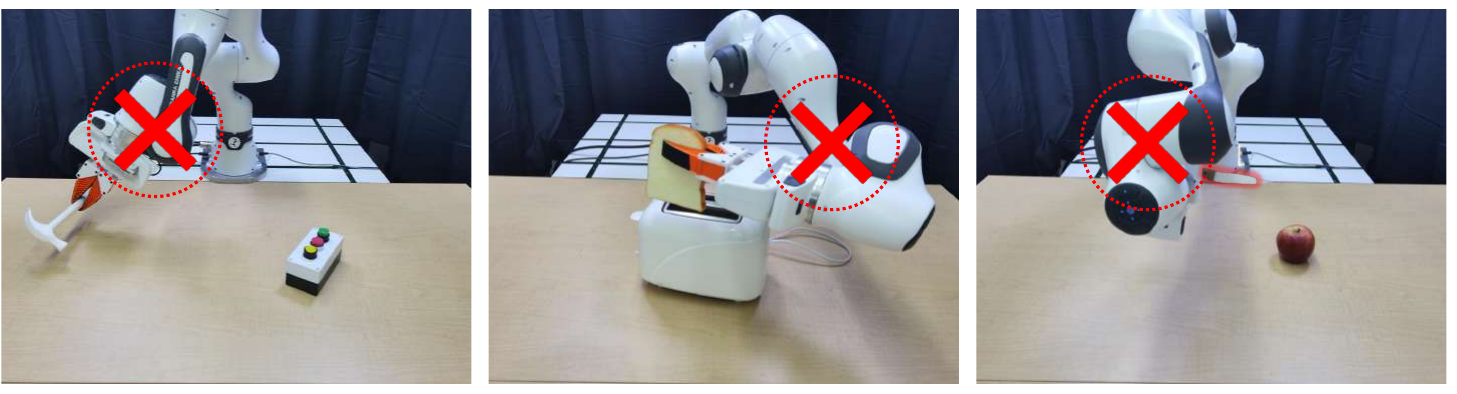}
    \caption{Example of IK failure.}
    \vspace{-0.5cm}
    \label{fig:example_ik_failure}
\end{figure}

\section{Environment Setup}
\label{appendix:environment_setup}

\begin{figure}[h]
    \centering
    \includegraphics[width=1\textwidth]{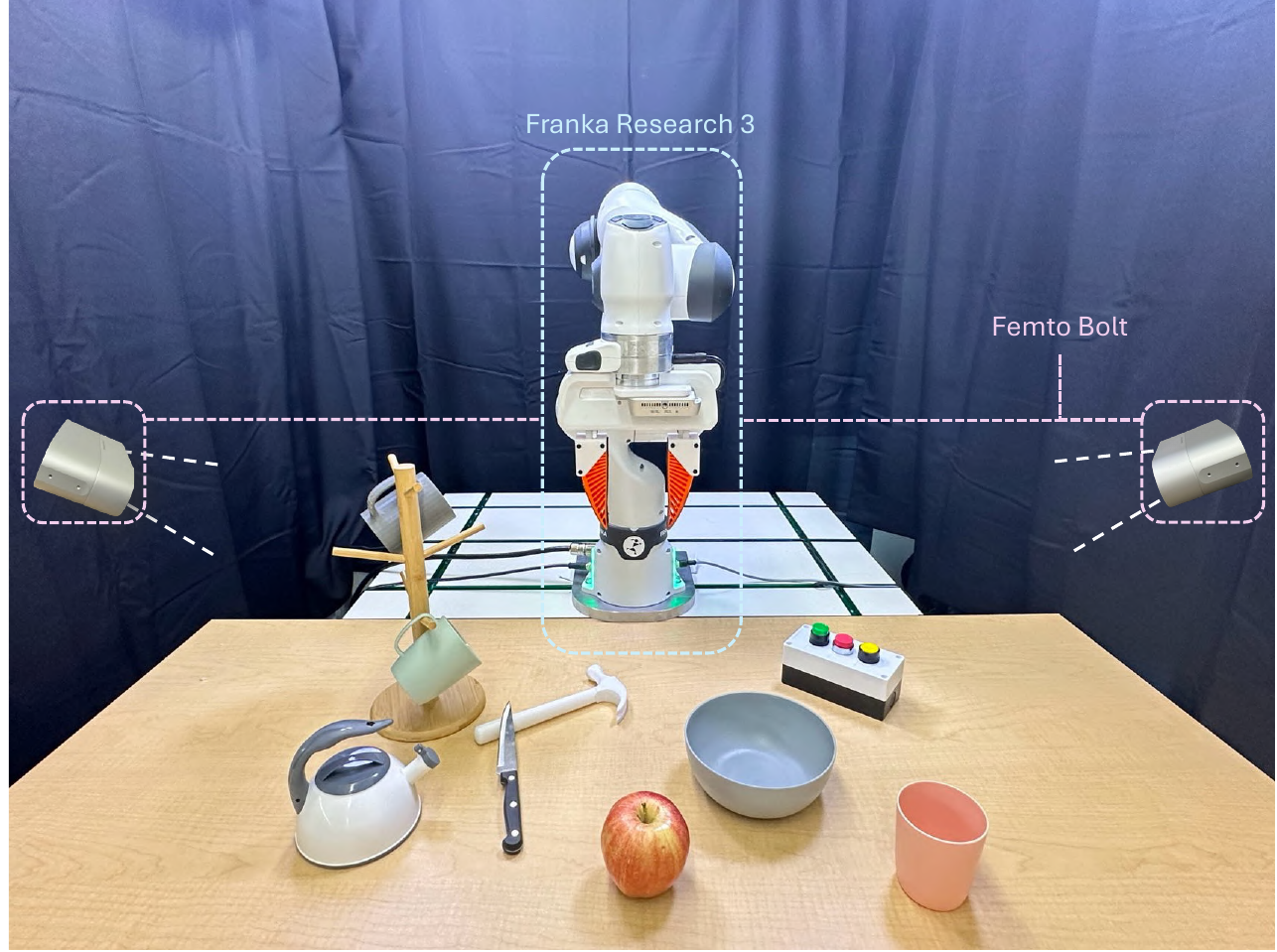}
    \caption{Real robot experiment setup.}
    \label{fig:example_ik_failure}
\end{figure}
Our task uses a Franka Research 3 robot, a 7-DOF manipulator. To capture comprehensive visual information and mitigate occlusions, we employ two Femto Bolt RGB-D cameras. These Time-of-Flight (ToF) based cameras are positioned on the left and right sides in front of the robot's workspace, providing robust depth perception across the scene. 

We utilize a diverse set of objects for our challenging object-to-object interaction tasks. These tasks require the model to not only identify the correct manipulation point on the source object but also to understand the intended interaction point on the target object. This necessitates a deep understanding of object-to-object affordance. Our tasks include:

\begin{itemize}
    \item \textbf{Pouring water from a teapot into a bowl or:}
    In this task, the functional affordance is the teapot's spout, which must be correctly aligned with the receiving area of the bowl (e.g., its bottom center) or the plate. The model must comprehend this specific object-to-object affordance for pouring. For instance, while the edge of the bowl might offer a grasping affordance, it is not the correct target for the pouring task. This task critically evaluates the model's ability to understand task-specific affordance.

    \item \textbf{Cutting fruit with a knife:}
    The model must identify the sharp edge of the knife (source object affordance) and the appropriate surface of the fruit (target object affordance) to perform a cutting action. This involves understanding the functional relationship between the knife's blade and the fruit's body.

    \item \textbf{Pressing a button with a hammer:}
    This task requires the model to recognize the head of the hammer as the striking surface and the button as the target. The interaction involves a precise contact between the hammer's affordance point and the button's surface.

    \item \textbf{Hanging a cup onto a mugtree:}
    The model needs to identify the handle of the cup (source object affordance) and a free peg on the mug tree (target object affordance). Successful execution depends on aligning the cup's handle to engage with the mugtree's peg.
\end{itemize}

These tasks are designed to test the model's understanding of object affordance and how they relate to each other in the context of specific actions.

\end{document}